# House Price Prediction Based On Deep Learning

Yuying Wu[1]  and Youshan Zhang[2]

## ABSTRACT

Since ancient times, what Chinese people have been pursuing is very simple, which is nothing more than "to live and work happily, to eat and dress comfortable". Today, more than 40 years after the reform and opening, people have basically solved the problem of food and clothing, and the urgent problem is housing. Nowadays, due to the storm of long-term rental apartment intermediary platforms such as eggshell, increasing the sense of insecurity of renters, as well as the urbanization in recent years and the scramble for people in major cities, this will make the future real estate market competition more intense. In order to better grasp the real estate price, let consumers buy a house reasonably, and provide a reference for the government to formulate policies, this paper summarizes the existing methods of house price prediction and proposes a house price prediction method based on mixed depth vision and text features.

The main research contents of this paper are as follows: firstly, it introduces the real estate related knowledge, analyzes the development status of the real estate market at home and abroad, and on the basis of previous studies, summarizes and analyzes two major types of factors affecting house prices, as the basis for the selection of data indicators, according to the indicators to determine the factors, it preliminarily selects 9200 sets of sample data, including image and text features, including from lianjia.com And Zillow, as well as the index data released by the National Bureau of Statistics and the four urban statistical bureaus of Beijing, Shanghai, Guangzhou, and Shenzhen. The economic factors include GDP, M2, GRP, CPI, and PPI, the population factor includes the total population of the region, and the time span is from 2017 to 2020. Secondly, this paper introduces several commonly used methods to predict house prices for comparison, summarizes the existing methods, finds that there are still some improvements, and puts forward some suggestions. In addition, the basic theory of deep learning is introduced in detail. One hot encoding and

[1] Liaoning University
[2] Lehigh University

resnet50 are used to preprocess the index attributes. The flow chart of the MVTs model is constructed, and the implementation steps of the proposed model are described in detail.

Finally, we select 75% of housing samples as training, and the remaining 25% of the samples as test. We compare the prediction results of our model with other five models: autoregressive integrated moving average mode (ARIMA), grey prediction model (GM(1,1)), support vector regression (SVR), BP neural network and artificial neural network (ANN), the results show that the proposed novel MVTs model has higher prediction accuracy. Therefore, MVTs is more suitable for the house price prediction.

# 摘 要

中国的人们自古以来追求的都很简单，无外乎"各安其居而乐其业，甘其食而美其服"，改革开放四十多年后的今天，人们已经基本解决温饱问题，亟待解决的就是住房问题。现如今由于蛋壳等长租公寓中介平台的暴雷增加了租房者的不安全感，以及近年来的城镇化和各大城市的抢人大战，这将使得未来的房地产市场竞争更加激烈。为了更好地把握房地产价格，让消费者合理购房，同时给政府制定政策提供参考，本文在总结现有房价预测方法的基础上，针对现有方法存在的不足，提出了基于混合深度视觉与文字特征的房价预测方法研究。

本文的主要研究内容为：首先介绍了房地产相关知识，分析了国内外房地产市场发展现状，并在前人研究的基础上总结分析了两大类影响房价的因素，作为数据指标选取的依据,根据指标确定因素中初步选取包括图像和文字特征在内的 9200 套样本数据，包括从链家网和 zillow 上收集到的关于房屋本体中影响房价的因素，以及国家统计局及北上广深四个城市统计局发布的各指标数据，金融因素包括 GDP，M2、地区生产总值 (GRP)、CPI、PPI，人口因素包括地区年末人口总数，时间跨度则从 2017 年至 2020 年；其次介绍了几种常用的后续会用于作对比的预测房价的方法，对已有方法的进行总结，发现仍有改进之处，提出 深度视觉与文本特征混合模型（MVTs）房价预测模型，此外详细介绍了深度学习的基本理论，运用 one-hot encoding 和 ResNet50 对指标属性进行预处理，构建了 MVTs 模型的实现流程图，并详细说明了提出的预测模型的实现步骤；

最后以选取的房屋样本数量的 75%进行样本训练，并以剩余的 25%的样本进行拟合和预测，并将预测的结果与自回归差分移动平均模 (ARIMA)、灰色预测模型(GM(1,1))、回归支持向量回归(SVR)、Back Propagation (BP) 神经网络和人工神经网络 (ANN) 5 种方法进行对比，结果显示提出的 MVTs 新模型能有更高的预测精度，因此它更适合用于房价的预测。

关键词：房价预测  MVTs  深度学习  图像识别  文字特征

# 目 录

# 图 表 目 录

## 图目录

## 表目录

# 绪 论

## 0.1 选题背景和意义

### 0.1.1 研究背景

房地产市场是每个国家国民经济的重要支柱，也是国家和地方经济发展的必要组成部分，为我国的 GDP 的增长做出了巨大贡献。2019 年中国的商品房销售面积同比增长 0.054%，但商品房的销售额增长了 6.76%，这说明经济的增长很大程度是通过房地产来拉动，这对于我国的经济转型和降低金融风险是没有好处的。现如今改革开放四十多年后的今天，人民生活水平不断提高，城镇化比例不断上升，农村人口不断的向城市迁移，必然导致住房需求的日益上涨，进而影响房价，同时房价的快速增长导致许多年轻人选择不结婚。在中国，房价每上涨 1%，初婚率下降了 0.31%[1]，类似的情况也可能同时发生在世界各地其他经济不断增长的城市中。从马斯洛的需求层次理论来看，"房子"给人类提供栖身之所，是安全感的来源，这是买房后的重要收获之一。房价工资上涨速度的不平衡，导致很多人买不起房，贫富差距日益突出，威胁着社会的稳定。中指研究院 2020 年 6 月《全国城市居民置业意愿调查报告》报告显示 5 月份到售楼处或中介门店看房高达 66%，其中一、二线城市购房需求非常高，占比 77%，并且由于蛋壳和青客等长租公寓中介平台的暴雷增加了租房者的不安全感，以及近年来的城镇化和各大城市的抢人大战，这将使得未来的房地产市场竞争更加激烈，有高达 51%的受访者的购房原因是预期未来房价会持续上涨。国内外有很多网站和中介平台为大众提供实时的房屋价格的预测和房屋推荐。如中国的链家和安居客和美国的 Zillow 和 Redfin 的出现会压缩部分中介作为以盈利为目的机构会为了获取高收益，信息的不对称使得购房者在预算内没有办法选择最优，承担经济损失的情况。但随着信息爆炸式增长，要从海量信息中选取有效安全信息，减少不必要损失，维护房地产交易市场的秩序以及合理健康发展。

仲量联行在 2020 年 10 月报告《智慧城市成功之路：人、房地产科技和房地产的有机融合》中揭示了房地产和科技的融合将成为未来城市发展的关键推动力。基于人工智能技术的进步所带来的房地产价格的预测结果准确性的提高，不仅为广大居民在进行购买和出售房屋时提供相关合理的参考，也有益于智慧城市的建设。美国的住房市场次贷危机所导致的一系列连锁反应就是一个反面教材，影响了全球经济的发展。因此，对房价进行高精度的预测是非常重要的。

## 0.1.2 研究意义

房价是体现经济运转好坏的重要指标，房地产开发商与购房者都密切关注着房价波动，构建有效的房价预测模型对金融市场、民情民生有着重要意义。对一个普通家庭来说，房子可以算得的上是最大家庭资产投入之一了，家庭的整体价值受房价波动的影响很大。房价的精准预测不仅对个人购房者是有利的，同时对于房屋出售者而言，可以直观了解当前房屋真实价格，防止要价过高造成不必要的损失。对比售房信息网站中介内的挂牌价格和实际售价，我们不难看出挂牌价格会比实际售价高，信息的不对称，就可能会让一些机构为了获得更高的报酬而哄抬物价有机可乘，导致房地产泡沫过高现象。现如今随着互联网大数据统计技术的发展，出现了如链家网，安居客，房天下等互联网平台使得现如今的房屋交易市场相对公开透明。信息爆炸式增长的今天，和股市一样，网络上充斥着大量的看涨或看跌的言论，他们或专业，或盲从，影响着人们的决策方向。那么此时，一个从房屋实际情况出发，考虑到多个影响因素在内的专业的，精准的房屋预测价格，就会在很大程度上更直观地分析房地产市场准确把握相关信息，从而根据自身需求及经济条件做出相关合理的决策。成为政府部门在对制定关于房地产或其他相关政策时的参考，采取针对性强，高效益的举措，让房价变得更加合理，进而使房地产市场的发展更加稳定，还能有助于预测通货膨胀和产出。对于银行等金融机构也可以根据政府政策方向，做出对自己有利的决策。因此，高精度地预测房价是有必要的。

然而，有太多影响房价的因素，这给房价预测带来了巨大的挑战。房价与其他一些宏观经济因素之间存在着微妙的相互作用，使得预测过程非常复杂。通过对国内外学者关于房屋价格的研究成果的梳理，我们发现大部分学者在影响因素上会从政治、社会、经济、人口等宏观层面对房屋价格进行预测，以及也有部分学者对建筑固有属性，如位置、卧室、浴室数量、房屋面积等微观因素影响下的房屋价格进行预测，在模型上通常会选择 GM(1,1)模型[43]、ARIMA 模型[3][48]、支持向量回归(SVR)[2][48]、时间序列预测[9][49]、神经网络[6][52][53][55]等方法进行预测。但上述的这些影响因素和方法都存在不足和局限性。此外大部分的文献都是对于单独某一个地区的房屋均价的预测研究，普适性不强，实用性稍有不足。

本文采用了深度神经网络模型预测了房屋价格，并研究了结合深度预训练模型特征和浅层卷积神经网络对房屋价格的影响。本文以中美两国各 4 个主要城市的房屋为研究对象，采用新的深度神经网络模型，最新的数据，大大提高了预测精度。

## 0.2 国内外文献综述

### 0.2.1 国内研究文献综述

研究针对房价预测问题，国内学者进行了大量的探索，通过对国内的房屋预测的方法的相关研究进行梳理，学者们提出的研究方法以及选取的指标皆有不同。由于房屋价格的预测是一个回归问题，学者们针对回归问题提出过许多解决回归问题的方法，例如 ARIMA 模型[3][48]、GM(1,1)模型[43]、支持向量回归(SVR)[2][48]、时间序列预测[9][49]、BP 神经网络[6][52][54]、人工神经网络[11][13]等。

早在上个世纪，就已经提出的了将 ARIMA 模型应用于城市房价预测。而后很对专家学者在原有模型上进行改进，提高模型的预测结果。侯普光和乔泽群[47]将小波分析与 ARIMA 模型相结合，通过对房价的数据进行分解和重构进行降噪，以及平稳性检验，估计参数而后建立相应的 ARIMA 模型进行预测。刘丽泽[8]则基于多元线性回归模型及 ARIMA 模型进行分析，对北京市的未来房地产走势进行预测，也针对房地产的行业发展提出来了建议。

欧阳建涛[42]首先提出将灰色预测模型应用在房地产价格预测，文中指出，灰色系统理论既可以较精确的描述过去和现在，也能在较少的样本数据下，采用方便的运算，准确的预测短期未来的房屋价格。而后李东月和马志胜[43]在原有 GM(1,1)模型的基础上进行了调整与改进，即连续选取近年份不同个数的连续的样本数据逐次建立相应的 GM(1,1)模型。发现了观察样本个数与误差之间存在一定关系，因此可以根据不同的研究问题的特点适当选取不同的样本个数，从而减少预测误差。刘琼芳[44]、盛宝柱和古玲[45]分别基于 GM(1,1)模型，对福州市和合肥市的房价进行预测。王莹和王志祥[46]则在 GM(1,1)模型的基础上，引入二阶弱化因子，对原始数据进行预处理，建立了更理想精度的模型预测了随后 5 年的淮安市的平均房价。

为了谋求算法思想上的创新以及构建新的模型框架，提出了使用支持向量机模型(SVM 模型),它具有高水平的小样本学习小样本学习能力，申瑞娜等[48]先用主成分分析的方法对初始数据降维处理，而后建立 SVM 模型，对上海的房价进行预测。结果发现预测精度较高，泛化能力较强。

武秀丽和张锋[9]采用时间序列分析法，以广州市几个有代表性的行政区的房价数据为分析对象来建立预测模型，并经过残差分析，对误差进行检验，结果预测值与实际观测值基本吻合，达到了预测的目的。而马尔科夫链将时间序列看作一个随机过程，利用概率建立一种随机型时序模型进行预测。谷秀娟和李超[49]、韦光兰[50]

等分别先将北京市和昆明市的房售价的月度统计数据进行时间序列的平稳化处理，对该城市房价未来走势进行预测，并对居民的消费及政府相关政策的出台提供参考建议。顾莹和夏乐天[51]也对马尔科夫链模型进行加权优化，具有一定的适用性。

在时间序列的基础上提出使用 LSTM 方法对房价进行预测，由于 LSTM 模型对时间序列具有良好的预测性能，Xiaochen Chen 等[57]人提出在原始 LSTM 模型上加以改进，使用 stateful LSTM 和 stack LSTM 模型对北上广深四个城市的平均房价进行高精度的预测，效果较比传统的 LSTM 好。

2006 年，加拿大多伦多大学教授、机器学习领域的泰斗 Geoffrey Hinton 和他的学生 Ruslan Salakhutdinov 在《科学》上发表了一篇文章，开启了深度学习在学术界和工业界的浪潮。现如今，需要大数据支撑的房价预测研究，神经网络就成为了大量学者采用的重要的预测手段之一，其中 BP 神经网络由于其本身在网络理论还是在性能方面已比较成熟，成为目前应用最多，最广泛的神经网络。高玉明和张仁津[6]基于遗传算法和 BP 神经网络进行发家预测，选取了 1998 年至 2011 年贵阳市的房价及其他影响因素作为实验数据进行试验，并与传统 BP 神经网络模型作对比，结果发现经过遗传算法优化后的 BP 神经网络预测模型收敛速度快，预测精度高。张卉[52]提出一种基于粒子群优化的 BP 神经网络算法的预测房价的模型，通过与传统 BP 神经网络的对比发现，新模型的收敛速度更快，预测精度更高。邱启荣等[54]为了提高房价预测精度，采用基于主成分分析的 BP 神经网络预测模型，仿真结果显示，对经过了主成分分析法重新组合新的综合指标的预测结果与原始值系统误差更小。验证了 BP 神经网络在放假预测中的有效性。王奕翔[55]则采用改进型 RF－BP 神经网络对房地产价格进行预测，避免了主观因素的影响，具有不错的预测效果。

闫妍等[53]根据 1998 年后季度房价数据样本量少的特点，以 TEI@I 方法为指导，并以小波神经网络矫正误差，分析得出政策和房地产投资之间存在抑制关系。

### 0.2.2 国外研究文献综述

国外学者也提出了一些不同的方法进行房价预测。当试图解释或预测房价时，我们需要修正房价的空间变化。一般来说，在所有可能的空间聚集中，如何在广阔的空间中选择一个空间细分并不清楚。Sommervoll 等[58]采用了一种受生物学启发的方法，即使用遗传算法来搜索具有良好的样本内和样本外性质的空间聚集，并根据它们在标准的 hedonic 房价模型中的解释力进行重组对房价进行预测。研究发现无论是在样本内还是样本外，遗传算法都能一致地找到优于传统聚集的聚集。同时对

不同运行的遗传算法的最佳集合的比较表明，即使它们收敛到类似的高解释力，它们在遗传和经济上往往是不同的。因此遗传算法一致地找到了样本内和样本外解释能力都很强的模型。

Montero[59]同样根据空间效应是房屋定价的内在因素的特点，提出了采用惩罚样条方法，建立了一个考虑了空间自相关、空间异质性以及光滑和非参数指定的非线性的参数和半参数空间特征模型变量的混合模型，根据一个包含西班牙马德里2010年第一季度的10512套住房价格和特征的数据进行预测，并将其与Hedonic房价模型研究结果做对比，研究结果表明，考虑空间异质性的非线性模型和房屋个别或区域特征与其价格之间的柔性非线性关系是房价预测的最佳策略。

Alfiyatin[60]利用回归分析与粒子群演算法，确定影响房价的因素房屋三个属性包括物理条件、概念和位置，具体包括邮政编码、房屋地址、经纬度、建造年份、建筑面积、土地面积、建筑价格（IDR/$m^2$）、土地价格（IDR/$m^2$）、距市中心距离（km）、校园数量、餐厅数量、卫生设施数量、游乐场数量、学校数量，传统市场或商场的数量，教堂数量，以及公共交通的便利程度在内的影响因素。PSO算法用于影响变量的选择，回归分析用于确定预测中的最优系数。研究结果表明，组合回归和粒子群算法是可行的，预测误差最小。粒子群算法的优点是在较小的搜索空间内PSO可以做最优解的预测，但搜索空间小也恰恰该方法的局限性。当数据含有较多的噪声时，会遇到过拟合现象。

在机器学习中，随机森林是一个包含多个决策树的分类器，预测方法后来也渐渐发展为基于树的模型方法。Rangan Gupta[61]为了最优地容纳考虑到的考虑了八个宏观经济因素，以及解释国家因子和预测因子之间的非线性关系，采用随机森林的机器学习方法。分析了宏观经济不确定性在预测美国各州和哥伦比亚特区房价同步变动中的作用。

Yang[62]提出了一种基于集成学习(ensembe-lerning)的房价预测模型，为了评估该模型的有效性，作者使用了额外树(extra trees)、随机森林(random forest)、GBDT算法和XGB算法来对比评估基准测试加州房价的效率。结果表明，与其它四种单一预测模型相比,采用集成学习(ensembe-lerning)的房价预测方法能提高预测精度和稳定性，同时改善了由于数据噪声的存在，而出现过拟合现象。

现如今，机器学习渗入到科学的各行各业中，同样有很多学者将神经网络应用于房价的预测，预测结果更精准，同时改善了过度拟合的问题。除了国内常用的BP神经网络外，国外学者针对房价预测提出了多种神经网络的组合和应用。

Limsombunchai[11]基于 ANN 模型，对新西兰克赖斯特彻奇的 200 个房屋信息进行预测，并将结果与经典房价预测模型——Hedonic 价格模型作对比，结果表明他们的 ANN 模型预测更精准。

Khamis[10]比较了多元线性回归（MLR）模型与神经网络模型在纽约房价预估上的表现。抽取 1047 套房屋样本，考虑到包括建筑面积、卧室数量、浴室数量、房屋占地面积和房屋建造年限对房屋的影响因素，进行模型比较。结果表明神经网络模型的均方误差（MSE）值也低于 MLR 模型，表明神经网络模型更适合作为房价预测的替代模型。但模型还是有一些局限性的，首先，二手房价格不是实际的销售价格，而是估计的价格。其次，这项研究仅考虑了特定年份的房屋信息，而忽略了时间影响。最后，房价可能会受到估计中未考虑在内的其他经济因素的影响。

Wang[13]设计了带有记忆电阻的多层前馈神经网络，实现了自动在线训练，通过 BP 算法可以进行调整人工神经网络的忆阻器的权重，以同步建立回归模型。使其能够在训练模式下学习预测房价，进而在预测模式下成功地预测房价通过美国波士顿几个城市的房价样本对神经网络进行训练和预测，预测结果接近目标数据。

Varma[12]等采用机器学习和神经网络的方法对房价进行预测，利用了线性回归、森林回归、增强回归进行系统优化，神经网络的应用进一步提高了算法的效率。作者同时还确定了影响房价的四个最重要的属性，分别是估价、类比房屋的销售价格、标价和浴室数量以及其他七个特征(房屋面积、卧室数量、浴室数量、地板类型、电梯情况、停车位置和家具状况)。并交叉验证了 14 种竞争模型的样本外预测性能，这有助于房价预测。根据所考虑的每个基本参数进行评估。结果表明，该方法比单独应用的算法具有最小误差和最大精度。

Serrano[64]基于 RNN 模型对时间序列数据，特别是价格进行预测。从房地产、股票和金融科技市场的领域进行了验证，实验结果表明，该方法能够对不同的投资组合做出准确的预测。

房价系统是一个多因素影响的复杂控制系统，受多种因素的综合作用，既有一定的规律性，同时又有随机性。尽管先前的研究分析了社会、经济、人口变化与房价之间的关系，但量化这种相关性不足以准确估计或预测房价。此外，以前的大多数工作都集中在房价预测的影响因素的文字特征上。Ahmed[2]等人在美国加利福尼亚州收集了 535 个样本房屋数据，其中既包含视觉图像又包含文本特征，他们是将视觉特征与文本特征组合进行预测的首次创新者。而后，他们应用所提取的特征，比较了 NN 模型和 SVR 模型在预测房价方面的效果，他们发现使用 NN 可以比 SVR

模型取得更好的结果。以及，与仅采用文本特征进行分析的结果相比，视觉和文本特征的组合产生了更好的预测精度。Zona Kostic[65]等人延续了将图像特征应用于在房地产价格预测建模中的必要性使用，选取了所有与属性相关的图像为基础，采用包括室内、室外和卫星视角在内的不同视角，论证了深度图像特征可以用来量化内部特征，并有助于模型的提高预测效果。甚至不需要通过任何人类参与，提出的技术就可以有效地描述可见特征，从而将其作为一种定量的度量引入到房屋预测建模中。但是，他们选取的样本量有限且影响因素的文本特征数量较少，数据所涉及得到的城市较少，忽略了城市之间的差异性，可能会导致过拟合。

## 0.3 本文研究内容及方法

针对以上文献的经验和不足的总结，提出了一个深度视觉与文本特征混合的房价预测模型(MVTs)，模型中有包括三个模块，分别为：深度特征模块、浅层 CNN 特征模块和文本特征模块，用于预测房价，以及两个不同的损失函数组合而成的新的损失函数，有效地度量了预测房价与实际房价的平均差。数据集方面，对应模型提供了一个大规模的房价预测数据集，它结合了深层的视觉和多种影响房价的因素特征，并建立一个单元的深度神经网络框架来解决上述模型中的不足，通过一些预先训练的深度神经网络，考虑深度视觉和文字特征来解决房价预测问题。

## 0.4 本文结构

首先是绪论部分，主要介绍了研究背景和研究意义，综合整理国内外文献综述，提出本文研究问题和方法。

第一章首先是相关概念的界定，包括房价，深度学习以及图像识别。其次阐述了房地产相关理论，介绍了影响中国房地产价格的因素，并分别分析了影响因素的作用机理，作为数据指标选取的依据。

第二章详细介绍了常被用于房价预测的模型，包括时间序列预测、GM(1,1)模型、支持向量回归(SVR)、BP 神经网络、人工神经网络。这些模型将被用来和本文章提出的模型的结果进行对比。

第三章根据指标确定因素中初步选取包括图像和影响因素特征在内的 8000 套样本数据，包括从链家网和 zillow 上收集到的关于房屋本体中影响房价的因素，以及国家统计局及北上广深四个城市统计局发布的各指标数据，金融因素包括 GDP，

M2、地区生产总值、CPI、PPI，人口因素包括地区年末人口总数，时间跨度则从2017年至2020年，并对收集到的数据预处理使其能够用于模型中。

第四章介绍了本文中提出混合深度视觉与文本特征的房价预测模型(MVTs)，构建了 MVTs 模型的实现流程图，并详细说明了提出的预测模型的实现步骤；

第五章以选取的房屋样本数量的75%进行样本训练，并以剩余的25%的样本进行拟合和预测，并将预测的结果与 ARIMA、GM(1,1)、SVR、BP、ANN 神经网络等方法进行对比，结果显示提出的混合深度视觉与文本特征的房价预测模型(MVTs)有更高的预测精度。

第六章对整篇论文进行总结，总结了自己的创新点和不足。对后续研究提供方向，并做出展望。

# 0.5 本文创新点和不足

## 0.5.1 创新点

1、提供了一个大规模的房价数据集，包括社会和房屋本身影响因素，结合了深层视觉图像和文本特征，该数据集在未来研究中可能具有重要价值。

2、首先提出一个单元深度神经网络的框架，将深度特征，浅层 CNN 特征和文本特征结合起来形成一个整体进行实验。实验表明，设计的体系结构可以以较低的平均绝对百分比误差输出房价。模型中最后提出了较为新颖的绝对平均差异损失函数，以确保平均预测价格更接近实际平均房价。

## 0.5.2 不足

对于数据集，我们提取到的房价预测的国内的数据集仅有北上广深四个城市共计 4000 套房屋样本数据，为了保证后续实验的准确性，我们仍需获得更多的数据。其次，房价的影响因素也不仅仅是上述文章中所搜集到的文本特征，仍有包括文化、政策，甚至是周边的配套设施情况，如学区，地铁等，仍有待搜集，但是相信，如果有相关的数据源，那我们的模型的准确度会更高。

我们的模型也有一些局限性。首先，测试数据的最低 MAPE 较高，这意味着预测的房价在某种程度上偏离实际价格。其次，两种视觉特征(深度特征和浅层 CNN 特征)的 MAPE 值都相对较低，说明图像的深层结构需要进一步研究。

# 1 相关理论基础

## 1.1 概念界定

### 1.1.1 房价定义

"各安其居而乐其业，甘其食而美其服"，一个满意的居住环境，会方便人类的生活，愉悦人的心情，充沛的精力会提高工作效率，投入到再生产活动中。因此房地产行业在我们中国这种以人为本的国家的整体经济规模中具有举足轻重的地位，是我国的支柱性产业。李克强指出，保障性住房建设是重大民生工程。全社会消费中对于房屋的消费权重相应较大，在很大程度上决定了整个市场消费水平。我国是计划经济体制，价格作为市场经济最重要的调节机制，调节房地产市场供求总量和结构。房价的合理波动对调节居民的生活水平有重要的功能和作用。

房价（房地产价格）是一个十分复杂的经济范畴，指建筑物连同其占用土地在特定时间段内房产的市场价值，即房价=土地价格+建筑物价格，房与地是不可分割的统一体，房价指的是这个统一体的价格。土地价格包括土地所有权价格（在中国特指对农村集体土地征用时的补偿价格）、土地使用权出让和转让价格，具体的土地价格根据房地产开发土地取得方式的不同，则形式不同。建筑物成本包括建筑过程中所有消耗的人力成本和建筑材料成本，具体包括开发商委托设计，监管，审计等专业机构工作以及施工前所必须班里的各项手续所花费的前期工程费；进行城市基础设施和公共建设施而必须投入的配套费；支付建筑承包商的建筑工程安装费；支付管理和销售人员的费用；以及贷款利息和税费等。

根据价格形成的方式不同，房价可以划分为：理论价格、评估价格和实际成交价格三种类型。理论价格又称基础价格，是指房地产内在价值的货币表现。评估价格又称参照价格，是指专业的房地产评估人员根据科学的方法对房子的市值估算出的价格。实际成交价格也就是市场价格，是指房地产交易双方实际的成交价格，其中受到供求关系、竞争烈度等因素的深刻影响。

房地产价格是一个十分复杂的概念。分析房地产价格的形式，是进一步研究房地产价格影响因素的基础，有利于制定合理的价格政策，采取科学有效的价格管理手段，保证房地产业的稳步发展。具体影响房价的因素也将会在这一章接下来的小节中进行详细描述。

### 1.1.2 深度学习

人们每天都会感知到大量数据，大脑也总是能够轻易的捕捉到有效信息，人工智能的核心就是做到对大脑的模仿，高效准确的捕捉到有效信息且能够表达。自 20 世纪 80 年代以来，机器学习作为人工智能领域的一门重要研究领域被提出，并在算法，理论和应用方面取得了巨大的成就。

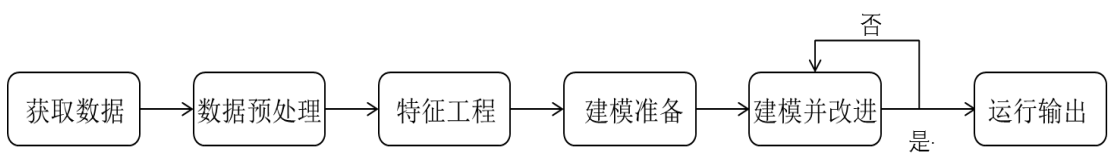

图 1-1 机器学习流程图

如图 1-1 所示，是机器学习在解决实际问题是的大致流程。首先通过爬虫，数据库，公开数据集等经由传感器来获取数据。而后对得到的数据进行预处理，对其进行清洗，特征转换，归一化等处理，以便于后续对特征的构建和提取。由于机器学习是数据和模型的结合，接着就是建模以及对模型的改进，若表现不好则需重新调试，直至运行表现良好，但仍需要对模型检测和更新，不断完善实验结果。系统中最重要的一环就是良好的特征表达，数据和特征决定了机器学习的上限，决定着最终算法结果的准确性。但之前的特征提取都是靠人完成的，浪费着大量的人力物力和时间，也不一定能够保证选取的质量。深度学习就解决了这一不足，特征工程能够自动化，不需要人的参与，这也是深度学习能被广泛使用的原因之一。深度学习作为机器学习中继浅层学习之后的又一次浪潮，克服了浅层学习的局限性，并且拥有更强的表示能力。

对于深度学习来说如图 1-2，其思想就是对堆叠多个层，来模拟人脑的分层结构，将上一层的输出作为下一层的输入，通过这种方式,把学习到的知识向下传递，就可以实现了对输入信息的分级表达。现如今也被广泛应用于声音识别，图像识别，文本数据的学习，进行预测。

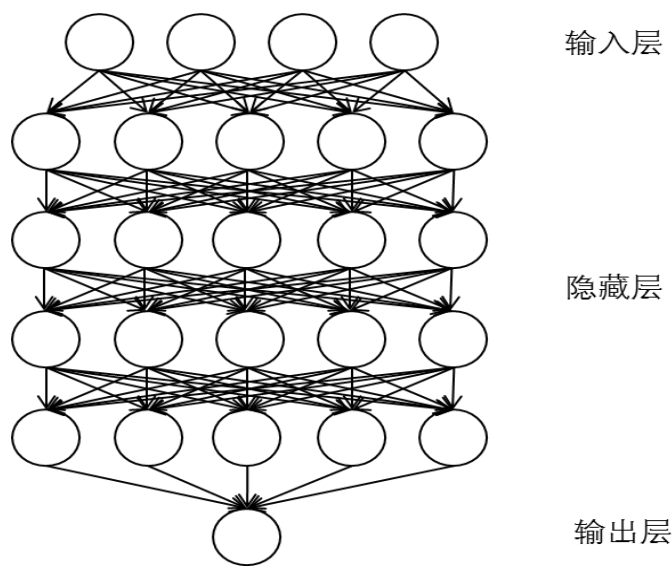

图 1-2 深度学习示意图

### 1.1.3 图像识别

上一小节介绍了深度学习的原理、优缺点以及实际应用。图像识别技术指的是利用计算机对图像进行处理、分析和理解，以识别各种不同模式的目标和对象，是最早被尝试应用的领域，也在深度学习的方法中得到了改善。

传统的图像识别方法模型层次较浅，也不需要依赖大量的图像为基础就可以快速的完成图像的识别。但这样会导致方法无法从原始图像中提取到更高深层的特征，且受人为因素影响，会导致图像识别率较低。由于深度学习方法其突出的特征提取能力和视觉特征捕捉能力，能够提取到更深层次的图像特征，且无人为或外界环境因素的干扰，使得图像提取特征表达性更强，准确率更高。现如今在人脸识别，医学图像识别，遥感卫星图像识别等方面取得了相当的成果。图像识别的第一步就是获取图像，这些图像可能是来自现实生活中的景物，人物，文字等通过相机或扫描方式采集，也可能卫星遥感或视频监控保存的一帧一帧的图像。而后提取到图像的深层特征应用于实际的模型中解决问题。

本文应用了深度学习中的图像识别，从每套房屋的客厅，厨房，卫生间，卧室以及外观的图片中提取深层特征。通过上述角度的图片能给购房者一个更加直观的展示，绿化、配套设施、装修等情况都有一个更加直观的认识，甚至可以考虑对家具进行更换、对装修材料进行调整等，有助于提高模型预测的准确度。

## 1.2 房地产经济学相关理论

为了解释房地产经济运行的理论，系统描述房地产经济运行过程，揭示房地产经纪运行规律提出了房地产经济学。房地产经济学是多学科的交汇，涉及到了多而复杂的理论，大体可分为相互关联相互依从的三个层面：

一是核心理论层如地租理论、房价理论、区位理论等；二是直接支配房地产运行的一般理论或内层理论，如房地产投资、房地产开发建设、房地产市场等理论；三是间接影响房地产经济运行的外延交叉理论。其中房价理论研究的房屋价格形成和变动规律的理论，房价理论的重点，是城市房价理论。当前城市房价理论包括房价基本理论、特征价格理论和收益模型理论。下面将针对房价理论进行详细描述。

### 1.2.1 房价基本理论

华人经济学家汪林海是房价基本理论的奠基人。经济学中，把商品所具有的满足人某种需要的能力叫做商品的有用性，房价基本理论中，住房的有用性指消费者对住房的大小、装修、地段等各种因素的综合评价，不考虑价格的情况下，购房者购买住房时，会尽可能选择有用性高的住房。劣等住房是指建在最差的地段上的住房，一般情况下，劣等住房就是城市边缘的住房。

汪林海认为，城市中某套住房的价格受到三个因素的影响：该套住房的有用性，城市边缘住房（劣等住房）的有用性，城市边缘住房（劣等住房）的生产成本。当在房产市场完全竞争时，城市中的某套住房的价格P与该套住房的有用性 U 以及劣等住房的生产成本 U0 呈正相关关系，与劣等住房的有用性$P_0$呈负相关关系。房价基本理论用一个公式来表示，就是：

$$P = \frac{U}{U_0} P_0$$

其中，P 为某套住房的价格，U 为该套住房的有用性，$U_0$为劣等住房（城市边缘的住房）的有用性，$P_0$为城市边缘的住房的生产成本。

### 1.2.2 Hedonic 房价理论

经济学中针对商品价格的研究是多种多样的，不同于传统的价格理论分析的商品是同质的，Hedonic 价格理论研究的是具有差异化的商品。Griliches 利用 hedonic 价格模型建立了汽车行业的价格指数。Ridker 和 Henning 最早开辟了一条将 Hedonic

房价理论应用于房地产市场的思路。而后以 Lancaster 的消费者理论和 Rosen 的特征价格理论为理论基础的特征价格模型（Hedonic Price Model），是近几十年来房价模型的主流模型之一。以 Kain and Quigley[68]，Wabe [69]，Anderson 和 Crocker[70]等为代表。

特征房价理论认为：第一、房屋本身对于购房者而言不会直接产生效用，而是经由房屋的各项特征而获得效用。第二、房屋本身之间的价格差异来源于住房之间的特征的差异[71]。第三、由于房屋不同的特征组成的数量及组合方式不同，使得房屋的个体价格产生差异。第四、住房由于具有各种满足购房者需要的特征，从而能按照某一价格出售，特征价格模型没有统一的理论定式，一般根据实际问题以及具体数据来确定，从而，在使用特征价格模型对影响住宅价格的因素进行研究时，可把住房的特征作为解释变量，把住房的价格作为被解释变量，用计量经济学模型来描述，就是：

$$P = f(C_1, C_2, C_3, \ldots, C_i) + \varepsilon$$

上式中，P 为某住房的价格，$C_i$ 为住房的特征向量（住房的面积，装修，位置，交通，学区等量化特征），ε 为随机误差项。

### 1.2.3 收益价格模型理论

在会计学领域的研究中,通常会这样一个问题,价格模型还是收益模型？其中价格模型具有更高的相关性，具有更强的经济解释;而收益模型的计量回归效果更好，更加可靠。收益模型理论认为房价是住房预期收益（即房租）的折现。收益模型为：

$$P = \frac{R_1}{1+t} + \frac{R_2}{(1+t)^2} + \frac{R_3}{(1+t)^3} + \cdots$$

其中，$P$ 为房价，$R_i$ 为历年的房租，t 为利息率。

### 1.2.4 房地产周期理论

由于房地产市场的发展受到多种包括人口，政治，宏观经济以及国家政策的影响因素，以及其自身运行规律的影响，会出现其发展的波峰和波谷期，而这一周期性的波动规律被称为房地产周期。任泽平任博士《房地产周期》书中提炼出一个精简的房地产周期理论，即"长期看人口、中期看土地、短期看金融政策"。其中人口、金融均属于需求侧因素，土地则属于供给侧因素，人口、土地、金融等综合决定了房地产周期。

#### 1.2.4.1 人口因素

人口、金融因素从需求侧角度来影响房地产周期。从根本上说，住宅最终是用于人居住用的，而且是人最基本的需求。人口总数、人口平均年龄、各年龄段的人口分布比例、人口的增长情况以及家庭结构都会影响到对住宅的需求，其影响房地产市场的逻辑是：

当人口红利和城市人口数量增加时，此时处于房地产周期的左侧，居民收入增长，可支配收入增加，提高了储蓄率以及消费能力的增长，带动了首次置业（20-35岁）和改善型置业（35-50岁）人群的购房需求的提高。同时外汇占款不断扩大，流动性过剩，使得房地产价格提高。人口红利逐渐消失，刘易斯拐点出现，房地产经济周期波动至右侧，经济增速减缓，居民收入涨幅不大，置业人数达到巅峰，住房逐渐饱和，迎来房地产黄金时代的结束，后房地产时代的出现"总量放缓、结构分化"的特征，人口迁移边际上决定房地产周期的进程。

#### 1.2.4.2 金融因素

房地产短周期是指在长周期趋势背景下，由于利率、抵押贷首付比、税率等短期变量引发的变化，改变了居民的支付能力和预期，使得购房支出提前或暂缓。影响房屋价格金融因素的影响机制如下：

国内生产总值 GDP 作为衡量宏观经济发展状况的指标之一，反映了总体经济状况。经济增长则投资增加，生产活跃，进而带动对房屋的需求，使得地价上涨；同时还会增加就业、工资上涨使得住宅需求增加。

广义货币 M2，包括流通中的现金和银行储蓄存款，反映现实和潜在的购买力，当 M2 增速都高于 GDP 增速，这必然会导致物价上涨，房价也会受到影响。

居民收入水平不同，影响对于房屋的需求。当人们从低收入上涨至高收入时，潜在的购房需求可以付诸行动。投资、投机需求也增加，从而使房价上涨。

物价水平反映的是整个经济的物价水平。一般选取 CPI 和 PPI 作为反映物价变动的指标。

CPI 即消费者价格指数能在一定程度上反映通胀水平，PPI 生产者价格指数，能够衡量工业品出厂价格变动程度。对于房地产市场，CPI 能间接反映商品住宅建造中的人工费变动情况，CPI 上涨则人工费上涨，同时使得购买力下降；PPI 能够反映材料费、机械费的价格变化水平，从而反映了住宅建造成本的变化。

利率影响了房地产企业的开发成本，贷款利率上升，所需支付的利息增加，成

本增加，房价上涨。对于购房者而言，抵押贷首府比上升，购房门槛提高，从而使住房需求减少，在当今供大于求的情况下，利率上升，则房价下降，反之则上升。

一组完整的房地产短周期为：当下调利率和抵押贷首付比时，居民支出水平提高，房地产销量增加，房地产开发商去库存，供给小于需求，致使房价上涨，房屋作为抵押物的价值上涨，会提高居民、开发商的贷款行为；当房价上涨至出现了泡沫化，政策上调利率和抵押贷首付比，居民支付水平下降，房地产市场进入淡市，商品房库存增加，供给大于需求，房价下降，其作为抵押物的价值缩水，会减少居民、开发商的贷款行为。在这个过程中，情绪加速器、抵押物信贷加速器等会影响房地产短周期波动。

#### 1.2.4.3 土地因素

土地是住宅的载体，从供给侧因素角度来看，政府的土地供给量从源头上决定了房屋的供给。根据国内外普遍经验表明，土地供给对房地产市场波动有显著影响。由于土地的供给缺乏弹性，自然土地的稀缺性以及政府的垄断行为使得可供人类利用的土地总量是刚性的，供给难以自动调节。德国房价的长期保持稳定，波动幅度较小,关键在于该国住房供给稳定且充足。日本在 1985-1991 年由于土地投机过度、供给不足且日本政府对此没有有效干预，导致房地产泡沫而后破裂，致使日本经济迅速倒退 30 年。在国内，房地产开发周期长，项目体量大，从取得土地开发资格到开工，再到预售或竣工待售项目周期一般为 1-3 年，供给存在滞后性。因此当住房市场上价格上涨时，开发商也难在短期向市场推售大量楼盘和住宅，反之当楼盘滞销价格下行时，开发商只可能放慢开发在建项目的节奏，也不会完全终止，永远滞后于市场的变化。事实上房地产周期波动的原因之一便在于供给的滞后性，并且土地供给政策还可能通过预期传导直接影响当期房地产市场。因此，土地因素主要在中期对房地产周期产生影响，介于人口因素和金融因素之间。

### 1.2.5 预期效应理论

预期效应是指人类的行为不是受他们的直接结果影响，而是受他们预期结果将会导致的结果所支配。在经济学中，当经济主体参与经济活动时，"预期"作为一种影响很大的因素往往会对决策者的选择和判断的产生影响。如果实际与预期相符，将加强预期的作用力和可信度，如果预期良好但实际上却差强人意，就会对人产生认知上的失调。

预期的不同会对购房者和开发商这两种不同的经济主体的行为产生相反的影响，对于购房者来说，会因为对房价的涨价预期而恐慌性购房，如果预期未来房价会下降，则会选择观望市场，持有货币等待，市场对于房屋的需求量就会下降。对于地产开发商来说，如果通过对市场研究后对未来市场的预期持积极态度，充满信心，形成了良好的前景预期，就会加大投入，增加供给，反之则减少供给。

经济学里也经常用"羊群效应"来描述经济个体的从众心理，在人们心中一直存在 "买房才能保值或升值，才能跑赢通货膨胀"的惯性思维，已经买了房的人自然不希望房价下跌，所以他们会更坚定这一想法，还没买房的人被不断催眠要买房直至成为催促别人买房人中的一份子。长此以往，陷入一个恶性循环当中，增加了需求，推高了房价。

## 1.3    本章小结

本章首先对相关概念进行界定，包括房价，深度学习以及图像识别。其次阐述了房地产相关理论，介绍了影响中国房地产价格因素，并分别分析了影响因素的作用机理，以作为指标选取的原则。

# 2 常用的房价预测模型

研究针对房价预测问题，国内学者进行了大量的探索，通过对国内的房屋预测的方法的相关研究进行梳理，学者们提出的研究方法以及选取的指标皆有不同。由于房屋价格的预测是一个回归问题，学者们针对回归问题提出过许多解决回归问题的方法，例如 时间序列预测[9][49][3][48]、GM(1,1)模型[43]、支持向量回归(SVR)[2][48]、BP 神经网络[6][41][52][54]、人工神经网络[11][13]等。

## 2.1 时间序列预测法

时间序列预测方法通过对历史数据分析处理，研究数据的基本特征和变化规律。该方法包括确定型和随机型两类。由于确定型时间序列预测方法容易忽略随机变量影响,而房价预测的变动有一定的特征，并在时间上有延续性，因此可以利用随机型时间序列预测方法做中短期预测。

随机型时间序列预测方法(Stochastic Time Series)STS 中存在不确定的固定协方差心$\delta_a^2(t)$与零均值，输出为预测值，可以分为以下几类：

### 2.1.1 自回归 AR (Autoregressive)过程

p 阶 AR (p)的公式如下：
$$y(t) = \phi_1 y(t-1) + \phi_2 y(t-2) + \cdots + \phi_p y(t-p) + a(t)$$
式中，y(t)为之前某些时刻的值和一个随机噪声的组合。

引入后移算子 B，有$y(t-1) = By(t)$，且$y(t-m) = B^m y(t)$，则公式 改为:
$$\phi(B)y(t) = a(t)$$
式中，$\phi(B) = 1 - \phi_1 B - \phi_2 B^2 - \cdots - \phi_p B^p$

### 2.1.2 移动平均 MA (Moving Average)过程

q 阶模型 MA (q)的表达式如下:
$$y(t) = a(t) - Q_1(t-1) - \cdots - Q_q a(t-q)$$

或$y(t) = Q(B)a(t)$

式中，$Q(B) = 1 - Q_1 B - Q_2 B^2 - \cdots - Q_q B^q$，a(t)一般由预测结果的残差或误差构造的，y(t)为随机噪声 a(t)当前值和过去值的组合。

### 2.1.3 自回归移动平均 ARMA (Autoregressive Moving Average)过程

由 AR(P)和 MA(q)组合形成的 ARMA(p, q)模型的公式如下：

$$y(t) = \phi_1 y(t-1) + \cdots + \phi_p y(t-p) + a(t) - Q a(t-1) - \cdots - Q_q a(t-q)$$

或 $\phi(B)y(t) = Q(B)a(t)$

### 2.1.4 自回归积分型移动模型 ARIMA

在引入差分算子 $\nabla$ 后一阶差分序列可写为 $\nabla y(t) = y(t) - y(t-1) = (1-B)y(t)$。d 阶差分序列改为 $\nabla^d y(t) = (1-B)^d y(t)$。经 d 次处理后，模型表达式为:

$$\phi(B)\nabla^d y(t) = Q(B)a(t)$$

在上述的模型中，其通用表达式为:

$$\sum_{i=0}^{p} \varphi_i Y(t-i) = \sum_{j=0}^{q} \theta_j a(t-j)$$

在常用的随机时间序列预测方法中，AR、MA 和 ARMA 都为平稳过程模型。由于房价波动有明显的周期性，因而属于非平稳的过程。实际上建立的模型是 ARMA 或 ARIMA 过程的模型。

随机型时间序列预测方法分为以下几个步骤：

1、平稳化原始数据，进行平稳性检验；

2、确定使用的模型，即通过原始数据的自相关函数与部分自相关函数确定适合使用哪种模型；

3、估计模型参数，建模预报模型；

4、检验预测模型精度与修正预测值。

通常来说，房价波动的时间序列预测都符合 ARMA(q, p)模型，但该模型求解的过程比较复杂。普光和乔泽群[47]将小波分析与 ARIMA 模型相结合，通过对房价的数据进行分解和重构进行降噪，以及平稳性检验，估计参数而后建立相应的 ARIMA 模型进行预测。

## 2.2 GM(1,1)模型

灰色系统理论首先是我国邓聚龙教授提出来的，部分信息未知且部分信息已知的系统就是灰色系统。我们对现实生活中的事物，不是一无所知，也不是无所不知。灰色系统的特点是先用少量的数据建立微分方程再根据此做预测，并将历史数据进

行累加,再对这个生成的数列建立起 GM (Grey Mode)模型称为灰色预测型，预测中常用的是 GM(1,1)模型,且数列有三种生成方式：累减生成、累加生成与级比生成。

灰色预测的本质是对已有的原始数据进行处理，以特定的数学模型模拟仿真这些数据，从而对将来某段时间的值进行预测。累减生成(Inverse Accumulated Generating Operation) $IAGO$ 与累加生成(Accumulated Generating Operation) $AGO$ 这两种方法是生成数列的方法。以 x(0) (k)表示采集数据数列中第 k 个数据，x$^{(r)}$(k)表示经过 r 次后得到的数列的第 k 个数据，得到 $AGO$ 公式如下：

$$x^r(k) = \sum_{m=1}^{1} x^{(r-1)}(m) = x^{(k)}(k-1) + x^{(r-1)}(k)$$

很明显，$AGO$ 与 $IAGO$ 互为逆运算，有: $x^{(0)} \xrightarrow{AGO} x^{(r)}, x^{(r)} \xrightarrow{IAGO} x^{(0)}$。但因为灰色理论所建立模型是生成数据的模型， GM (1,1)模型的预测值一般要作生成运算的逆运算，才能得到需要的预测结果。

灰色预测生成处理的另一个好处是利用较少的数据获取数据中存在的规律，与需要统计大量样本数据的算法相比，有更强的灵活性和适用性。

灰色预测采用的是 GM (1,1)的模型，两个 1 分别表示 GM 模型是 1 阶和 1 个变量的微分方程，这个变量指的是预测量。

GM (1,1)的微分模型为：

$$\frac{dx^{(r)}}{dt} + ax^{(r)} = u$$

式中，系统参数 a 和控制量 u 构成的待辨识参数列 $\hat{a}$ 为：

$$\hat{a} = \begin{bmatrix} a \\ u \end{bmatrix}$$

若只进行一次生成处理，即 r=1 时，微分模型为:

$$\frac{dx^{(1)}}{dt} + ax^{(1)} = u$$

进行差分处理后得:

$$a^{(1)}[x^{(1)}(k+1)] + aX^{(1)}(k+1) = u$$

式中：$a^{(1)}[x^{(1)}(k+1)] + aX^{(1)}(k+1) - x^{(1)}(k) = x^{(0)}(k+1)$

背景值：

$$X^{(k+1)} = \frac{1}{2}[x^{(1)}(k) + x^{(1)}(k+1)]$$

有：

$$k = 1, x^{(0)}(2) = a(-\frac{1}{2}(x^{(1)}(1) + x^{(1)}(2))) + u$$

$$k = 2,\ x^{(0)}(3) = a(-\frac{1}{2}(x^{(1)}(2) + x^{(1)}(3))) + u$$
$$\vdots \qquad \vdots$$
$$k = n,\ x^{(0)}(n+1) = a(-\frac{1}{2}(x^{(1)}(n) + x^{(1)}(n+1))) + u$$

令：

$$y_N = \begin{bmatrix} x^{(0)}(2) \\ x^{(0)}(3) \\ \vdots \\ x^{(0)}(n+1) \end{bmatrix},\ B = [X \vdots E] = \begin{bmatrix} -\frac{1}{2}(x^{(1)}(1) + x^{(1)}(2)) & 1 \\ -\frac{1}{2}(x^{(1)}(2) + x^{(1)}(3)) & 1 \\ \vdots & \vdots \\ -\frac{1}{2}(x^{(1)}(n) + x^{(1)}(n+1)) & 1 \end{bmatrix}$$

有 $y_N = B\hat{a}$，再由最小二乘法，得到$\hat{a}$模型最终的参数如下：

$$\hat{a} = (B^T B)^{-1} B^T y_N$$

A 与 u 是固定值，在建立模型过程中，要对建立的 GM(1,1)模型进行不同的检验。

一个完整的灰色预测流程应该包括以下步骤：

1、对原始数据进行生成处理，减少其随机性，增强其规律性；

2、在生成数据基础上，建立 GM(1,1)模型，用最小二乘法辨识参数$\hat{a}$；

3、对辨识后的模型检验，如果模型可信，进入下一步；否则，转到(1)；

4、 根据建立的 GM(1,1)模型预测将来的值；

5、将步骤(4)的预测结果作逆生成处理，就可以得到真实的预测值。

欧阳建涛[42]首先提出将灰色预测模型应用在房地产价格预测。由于灰色预测法有要求数据少、运算简单而且方便、中短期的预测精度高等特点，应用广泛，但仍一定局限性，例如对于长期的预测中预测值会有一定的偏差。

## 2.3 支持向量回归(SVR)

针对二进制分类问题，提出了原始的线性支持向量机模型(SVM)。给定数据及其标签：(xn , yn), n=1,…,N 和  yn∈ {−1, 1}。它旨在优化以下等式：

$$min_{w,b} \frac{1}{2}||w||^2 + \lambda \sum_n \xi_n^2\ s.t.$$
$$y_n(w^T x_n + b) \geq 1 - \xi_n (\forall_n),\qquad \xi_n \geq 0\ (\forall_n)$$

其中 λ 控制边距的宽度(较小的边距和较小的 λ)；$\xi_n$是一个非负的松弛变量，对与边距相对应的数据点进行惩罚； b 是偏差。线性 SVM 也可以用作回归方法（称为 SVR），在分类问题上与 SVM 相比几乎没有细微差异。首先，SVR 的输出是一

个连续数字，而不是分类问题中的类。此外，SVR 中存在容限 ε。但是，主要思想始终是相同的：将误差最小化并将利润最大化。它的目的是是优化以下约束条件：

$$min_{w,b} \frac{1}{2}||w||^2 + \lambda \sum_n (\xi_n + \xi_n^*) \, s.t. \, y_n - (w^T x_n + b) \leq \varepsilon + \xi_n$$

$$y_n - (w^T x_n + b) \leq \varepsilon + \xi_n^* (\forall_n), \xi_n, \xi_n^* \geq 0 \, (\forall_n)$$

其中 $\xi_n^*$ 是另一个非负的松弛变量[40]。

SVR 模型在线性函数两侧制造了一个"间隔带"，对于所有落入到间隔带内的样本，都不计算损失；只有间隔带之外的，才计入损失函数。之后再通过最小化间隔带的宽度与总损失来最优化模型。如下图这样，只有那些处于带边缘之外，或落在隔离带边缘上的样本才被计入最后的损失：

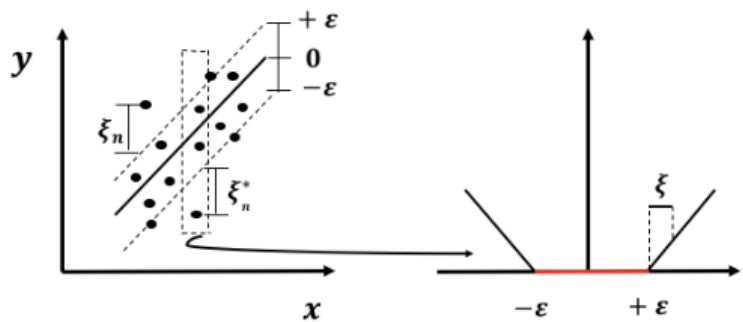

图 2-1 具有 ε 密集带的一维线性 SVR。虚线矩形可以在左侧图像中替代显示

## 2.4 神经网络

### 2.4.1 BP 神经网络

BP(Back Propagation）神经网络是 1986 年由 Rumelhart 和 McCelland 提出的，是一种应用广泛的多层前馈神经网络之一。输入层神经元是接收外界传递的消息，中间层负责信息的变换，由信息变化能力的需求，中间层能够确定是单隐层或者多隐层的结构，由输出层向外界输出处理后的结果。当实际输出与期望输出不令人满意的时候，就进入了误差的反向传播阶段。该阶段利用最速下降法，经过输出层，通过不断地调整反向传播网络的权值与阈值，使得训练网络误差的平方和最小，它是神经网络自主学习并训练的一个过程，这个过程会一直进行直到网络输出的特定的误差之下或设定的学习次数为止。BP 神经网络的结构图如图 2-2 所示：

假设输入的训练样本是 $X_k = (x_{k1}, x_{k2}, \cdots, x_{kh}, \cdots, x_{kn})$，那么实际的输出为 $Y_k = (y_{k1}, y_{k2}, \cdots, y_{kj}, \cdots, y_{km})$。

m 个分量对应输入层 m 个神经元，那么相应地 $X_k$ 的理想输出为：$Y'_k = (y'_{k1}, y'_{k2}, \cdots, y'_{kj}, \cdots, y'_{km})$。假设训练网络输出的误差函数是：

$$E_k = \frac{1}{2}\sum_{j=1}^{m}(y'_{kj} - y_{kj})^2$$

本文中的作用函数（又称为刺激函数）取的是（0,1）内连续的 Sigmoid 函数：$f(x) = \frac{1}{1+e}$，对于输入样本 $X_k$，对应输出层的实际输出为：$y_j = f(U_j)$

输出层单元 j 的输入为:

$$U_j = \sum_{i=1}^{p} w_{ij}O_i + \gamma_j$$

隐含层的实际输出为：$O_i = f(I_i)$，隐含层 i 的输入为：

$$I_i = \sum_{h=1}^{n} v_{hr}x_h + \theta_i$$

隐含层误差为：

$$e_i = \frac{1}{2}(1 - O_i^2)\sum_{j=1}^{m}\delta_j w_{ij}$$

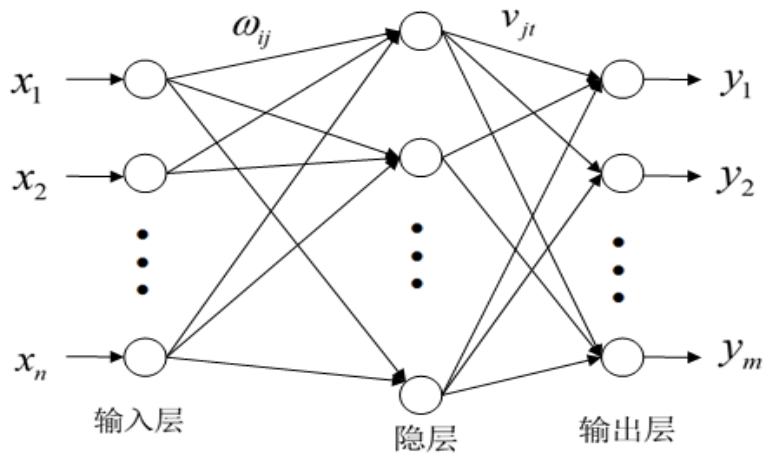

图 2-2 BP 神经网络

高玉明和张仁津[6]基于遗传算法和 BP 神经网络进行发家预测，选取了 1998 年至 2011 年贵阳市的房价及其他影响因素作为实验数据进行试验，并与传统 BP 神经网络模型作对比。

### 2.4.2 人工神经网络

人工神经网络法,因其有并行处理能力与高度非线性泛化,它并不需要事先确定函数的形式,而是先对采集的历史数据进行训练,得出影响电能耗各种因素与预测负荷的关系。它有记忆信息、推理知识、优化计算与自主学习的特点,它的自学能力与自适应功能是一般算法和专家系统不能比拟的。

人工神经网络的结构模型有递归神经网络(Recursive Neural Network,RNN)、自适应共振理论(ART)函数联接网络(Functional Link Network,FLN), GMDH 网络、自组织竞争(Kohonen)网络、双向联想记忆(BAM)网络、径向基函数((Radial Basis Function RBF)网络、Boltzmann 机、小脑模型关节控制器(Cerebella Model Articulation Controller, CMAC)、盒中脑(BSB). Hopfield 模型、CPN 模型等。其中感知器、BP 网络、RNN 网络、RBF 网络以及 Kohonen 网络等模型都有现成的研究成果。

但同时人工神经元网络也有一些缺点：

(1)人工神经元网络一般难以解释并给出实际的物理意义；

(2)需要估计的参数比一般的统计预测模型多,使网络的过度拟合现象,也比一般的统计模型耗费时间长；

(3)在进行非线性问题预测上,只解决了存在性问题,尤其是针对不同的问题,选择什么样的人工神经元网络结构,目前还没有成熟的理论；

(4)对网络的研究不能只停留在对学习方法的改进上，更应该深入研究网络映射机理的进行，只有找出网络模型在做非线性预测收敛速度慢的症结所在,并能够研究映射特点，才能够探索新网络的结构。

从总体上看，人工神经网络有下列的特点与不足：

特点：1、原理清晰、易于理解；2、充分考虑影响变化规律的因素；3、网络模型能够进行非线性模拟，结果与现实接近；

不足：1、样本选择较困难；2、网络的收敛速度慢；3、易陷入局部最小值。

Limsombunchai[11]基于 ANN 模型,对新西兰克赖斯特彻奇的 200 个房屋信息进行预测。Serrano[64]基于 RNN 模型对时间序列数据，特别是价格进行预测。从房地产、股票和金融科技市场的领域进行了验证，实验结果表明，该方法能够对不同的投资组合做出准确的预测。

# 3 数据指标说明及处理

## 3.1 数据来源

为了检验模型的普适性，本文的选择的数据是中美两国的房屋数据，并对比两个国家影响因素的区别和影响的程度的差异。通过阅读大量相关文献，对两国行情的不同，提取了不同的指标特征。国内的房屋样本数据，我们从链家网上的公开信息中手动随机收集了中国北京、上海、广州、深圳 4 个城市的 4000 套的房屋图像及其相关影响因素，以及国家统计局及北上广深四个城市统计局发布的各指标数据，金融因素包括 GDP，地区生产总值、M2、CPI、PPI，人口因素包括年末人口总数。我们还用同样的方式从 zillow.com 上搜集美国洛杉矶，华盛顿，纽约和波士顿 4 个城市的 4000 套房屋数据，同样包括文本特征和视觉图像特征。时间跨度为 2017 年至 2020 年，基本是现阶段的最新数据。

## 3.2 数据指标说明

本文的主要模型是将房屋影响因素特征与视觉图像特征相结合,并通过深度神经网络提取图片中的文本特征建立房价预测模型。

### 3.2.1 影响因素特征

通过之前的分析，我们知道了影响房价的因素包括社会因素和本身因素，我们选取社会因素中的金融因素包括 GDP，M2、地区生产总值、CPI、PPI，人口因素包括各地区年末人口总数，以及房子的本体指标。通过阅读相关文献，我们发现 Khamis[10]考虑到包括建筑面积、卧室数量、浴室数量、房屋占地面积和房屋建造年限对房屋的影响因素。Varma[12]确定了七个房屋固有特征——房屋面积、卧室数量、浴室数量、地板类型、电梯情况、停车位置和家具状况，以及三个最重要的属性，分别是估价、类似房屋的销售价格、标价。

国内的房屋的本体属性特征，我们根据同等环境下，普通购房者在进行选择时会考虑到的影响特征，选取了包括室、厅、厨、卫的数量、建筑面积（m2）、楼主体建造年份、楼房整体高度、房屋所在楼层、户型结构、建筑类型、建筑结构、是否有电梯、梯户比、装修情况、房屋朝向、房屋实际出售价格、挂牌价、邮编、地址、出售时间在内的 20 个特征。但获取的特征值并不总是连续值，而有可能是分类值。例如房屋所在楼层分为高、中、低；户型结构分为平层、复式、错层；建筑类

型分为板楼和塔楼；建筑结构分为钢混、砖混、和钢砖混合；以及是否有电梯、装修情况、房屋朝向。这些离散的特征，后续将通过数据预处理，使其易于利用。

从美国的房地产市场实际情况出发，国外的文本特征我们主要选择扩展收集以下主要因素，分别是卧室数量、浴室数量、房子的建筑面积、邮政编码、建造年份、重修年份、房屋类型、停车位和采光率。因此，我们对一所房屋共拥有包括房价在内的 10 个特征。他们有些并不是连续值，后续我们将通过同国内一样的通过 one-hot 编码方法，使其适用于机器学习，应用于我们的模型中。

### 3.2.2 视觉图像

通过阅读相关文献，最近，研究界对房地产估价问题表现出了兴趣，例如检查视觉因素的影响[Image Based Appraisal of Real Estate Properties]或为房地产得出明确的视觉风格[Vision-based Real Estate Price Estimation]。学术界已经开始将图片作为房地产估价过程中最重要的因素之一。Ahmed[2]等人收集到的房屋样本数据中既包含视觉图像又包含文字特征，与仅采用文本特征进行分析的结果相比，视觉和文本特征的组合产生了更好的预测精度，这一点在 Zona Kostic [65]的文献中也得到了验证。因此我们继续延续了视觉图像与文本信息相结合的特征选取。

中国由于庞大的人口数量，复杂的经济环境以及中国目前正处于的城市化阶段，虹吸效应促使越来越多的人涌入一、二线城市，北上广深由于本身经济发展快速，机会与机遇更多，使得城市承载更多的人口，况且未来大家不断地提升自我需求，提升独居率，房子盖的永远不够，所以国内大都以高楼大厦为主，因此我们选择的包括客厅，厨房，卧室和卫生间在内的四张图片为一组。

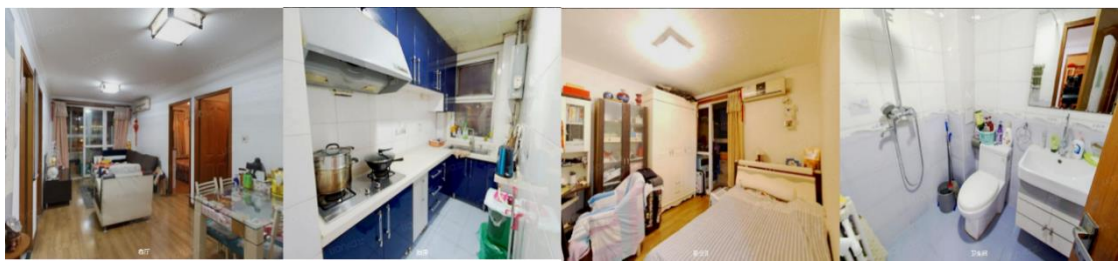

图 3-1 中国房屋视觉图像，由四个不同的视图表示：客厅、厨房、卧室和卫生间

由于国外的房屋大都独门独户，所以我们选取了如图所示的四个角度，他们分别是房屋的外观，浴室，卧室和厨房的图片。通过图片我们可以观察到很多主观影响因素，例如对于大部分人来说，从房屋的本体外观来看，房屋的高度，房屋外观材质状态，外墙体是否干净整洁；图片中房屋外的环境。卧室，浴室和厨房是每个房屋在人们的日常生活中必备的区域，不同的房间面积、数量，家具情况等都会产生不一样的预测结果。

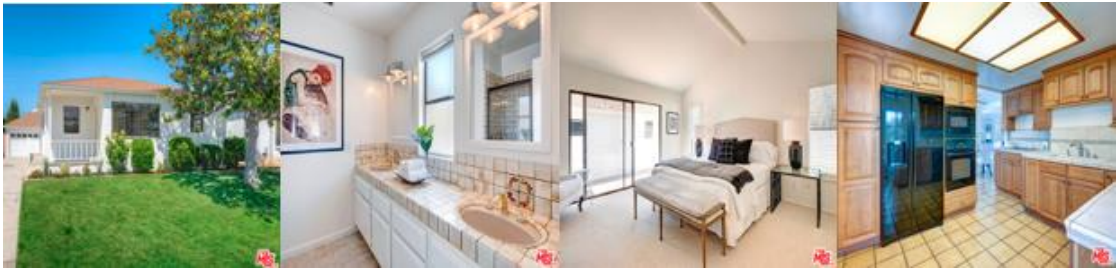

图 3-2 美国房屋视觉图像，由四个不同的视图表示：正面、浴室、卧室和厨房

## 3.3 数据预处理

由上一节的描述我们可以看出有些我们收集到的数据并不能直接用于模型的训练，有些不连续的或者缺失的特征值需要我们预先处理，使其适用于后续的模型学习和训练。同样，图片中也包含了影响预测效果的深度特征。

### 3.3.1 文字特征预处理

在实际的机器学习的应用任务中，特征有时候并不总是连续值，有可能是一些分类值，在机器学习任务中，对于这样的特征，通常我们需要对其进行特征数字化，如下面的例子：

有如下几个特征属性：

- 房屋所在楼层：      ["高", "中", "低"]
- 户型结构：          ["平层", "复式", "错层"]
- 建筑类型：          ["板楼". "塔楼"]
- 建筑结构：          ["钢混", "砖混", "钢砖混合"]
- 是否有电梯：        ["无", "有"]
- 装修情况：          ["简装", "精装", "其他"]
- 房屋朝向：          ["南", "北", "西", "东"]

对于某样本，["高"，"平层"，"板楼"，"钢砖混和"，"否"，"简装"，"西南"]和["中"，"复式"，"塔楼"，"钢混"，"是"，"精装"，"东北"]

我们需要将这个分类值的特征数字化，最直接的方法，我们可以采用序列化的方式分别将他们表达为[0,0,0,2,0,0,1,20]和[1,1,1,0,1,0,1,31]。但是这样的特征处理并不能直接放入机器学习算法中。

对于上述的问题，建筑类型和是否有电梯是二维的，房屋所在楼层、户型结构、建筑结构和装修情况是三维的，房屋朝向是多维的。这样，我们可以采用 One-Hot Encoding 的方式对上述的样本。

One-Hot Encoding，又称为一位有效编码，主要是采用 N 位状态寄存器来对 N 个状态进行编码，每个状态都由他独立的寄存器位，并且在任意时候只有一位有效。即保证每个样本中的单个特征只有 1 位处于状态 1，其他的都是 0。

自然状态码为：　　　000, 001, 010, 011, 100, 101

独热编码为：　　　000001, 000010, 000100, 001000, 010000, 100000

下图以户型结构为例，解释 One-Hot Encoding。它的优点：一是解决了分类器不好处理离散数据的问题，二是在一定程度上也起到了扩充特征的作用 。

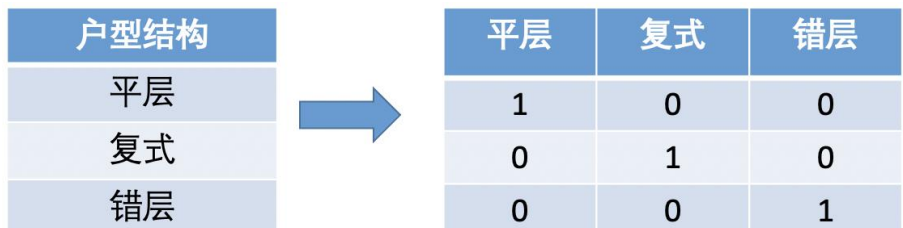

图 3-3 文本特征的 one-hot encoding

### 3.3.2 图片深度特征

为了提取深度图像特征，为了有效地组织输入数据，我们分别创建了一个由 4 个图像组成的有序序列分别包括厨房、浴室、卧室和客厅以及厨房、浴室、卧室和房屋外观，使用图 3-1、3-2 所示的顺序。

在深度特征模块中，我们遵循协议[14][27]，使用预训练的 ResNet50 模型提取的特征向量，从最后一个完全连接层提取所有特征。因此，一个图像的最终输出为一个 1×1000 的向量。特征提取分为两个步骤：1、将图像进行预训练的神经网络重新调节为不同的输入尺寸；2、从最后一个完全连通层提取特征。在特征提取之后，接下来两个重复的模块。在每个模块中，有一个稠密层和一个"ReLu"激活层。全链接

层的单位分别为 1000 和 4。CNN 专注于基于属性的细节，这些细节可以有效地利用不同房源之间的风格差异。在浅层 CNN 特征模块中，包括三个重复模块，每个模块有四层：卷积二维层、"ReLu"激活层、批量归一化层和最大池(max pooling)层。卷积二维层中三个模块的核大小为(3，3)，滤波器大小分别为 16，32，64，最大池(max pooling)层的池大小是(2，2)。在平坦层之后，是与深部特征模块相同的模块，但稠密层的单位是 16 和 4。

图 3-4 显示了房屋的视觉和文字特征以及使用预训练的 ResNet50 模型提取的特征向量。左上方是手动选取的原始房屋图像，右上方是从 ResNet50 模型中提取的特征，其中显示了特征出现的频率。在底部，我们列出了房屋的十个属性，分别是卧室数量，浴室数量，房屋面积(平方英尺)，邮政编码，建造年限，重修年限，房屋类型(0：单独住宅，1：联排别墅，2：高档公寓，3：多单元住宅，4：普通公寓)，停车位数量，采光率以及房屋价格(M：百万)[56]。

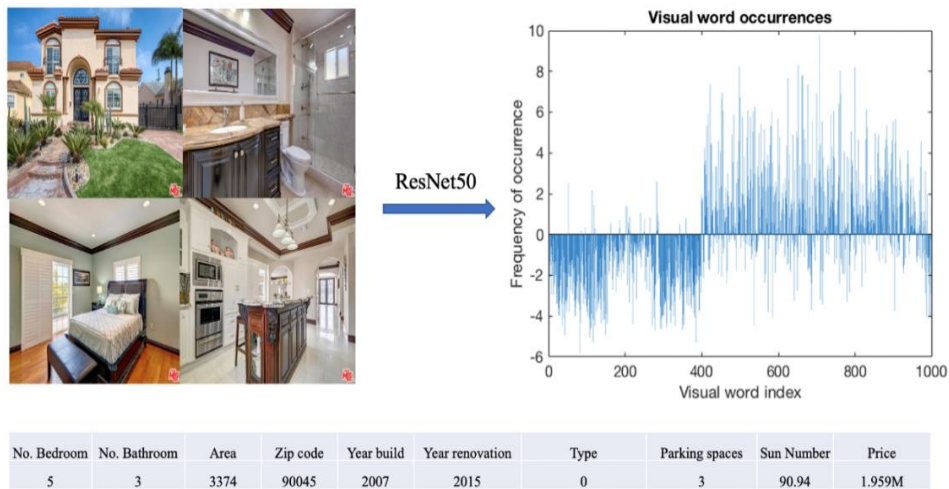

图 3-4 一所房子的所有样本特征

# 4 深度视觉与文本特征混合的房价预测模型(MVTs)

## 4.1 深度神经网络

目前,深度学习是机器学习的一个热门领域。深度学习是最有效的,有监督的,时间和成本效率最高的机器学习方法。为了开发各种智能世界系统,深度学习已经占据了工业和研究领域的主导地位。深度学习在大型复杂数据集近似化和简化为高度精确的预测和转换输出方面显示出巨大的潜力,极大地促进了以人为中心的智能系统。与为单一的不灵活任务开发的复杂硬编码程序不同,深度学习体系结构可以应用于所有类型的数据,无论是视觉、音频、数字、文本还是某些组合。此外,先进的深度学习平台正变得越来越复杂,通常是开源的,可供用户使用广泛使用。此外,包括谷歌、微软、亚马逊、苹果、百度、腾讯等在内的大公司都在大力投资深度学习技术,以提供能够进一步提高深度学习性能的硬件和软件创新,可用于下一代智能世界产品。

### 4.1.1 深度神经网络的层

简单的介绍一些常用的神经网络层。包括卷积层,池化层,分批归一化层,全连接层,脱落层和回归层,并以 AlexNet[29]做一个简单的例子。

1、卷积层(Convolution layer)

在卷积层中,在每个像素上传递一个特定维度的核或滤波器,也可以由指定要跳过的像素数的跨距值来确定。该跨步值指定要跳过的像素数。图 4-1 是一个 6×6 的卷积操作例子,每个 $X_i$ 代表着像素强度值,和尺寸为 3×3 的过滤器,每个像素的权重为 $W_i$,当滤波器以 $X_8$ 的值为特定像素中心时的卷积运算的公式。通过在整个图像上重复卷积运算将生成输出图像。在所有的神经网络中的卷积层执行类似的数学运算。输出图片的大小取决于过滤器、跨距和填充的大小,可由以下公式确定:

$$size\ of\ the\ output\ map = \frac{size\ of\ input - filtersize + 2(padding\ value))}{stride}$$

输出图片的深度取决于在输入上应用的滤波器通道的数量。

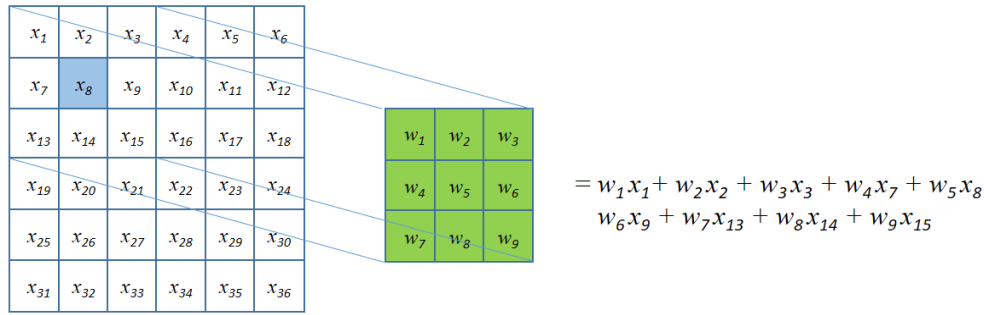

图 4-1 卷积运算

2、激活层(ReLU)

通过将卷积运算的结果图通过 ReLu 获得激活层。可以使用的激活功能有许多，例如 tansigmoid, ReLu, leaky ReLu 等。大量文献显示，ReLu 的使用最广泛。ReLU 函数可以用数学表示为:

$$f(x) = max(0, x)$$

3、池化层(Pooling layer)

在最大池层中，找到激活层上过滤器的接收字段中的最大值，并将其放置在相应的位置，而不丢失其空间信息。在图 4-2 中显示一个 4×4 的激活层，和一个 2×2 的过滤器的最大池化操作的示例。这减小了激活层的维度，也称为缩减采样。不同的深度神经学习网络有着不同的最大池和尺寸跨度，用于下采样。

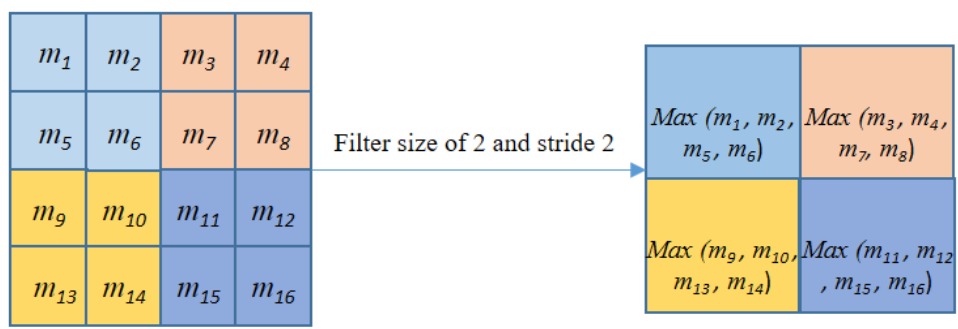

图 4-2 最大池化操作

4、跨渠道或分批归一化层(Cross channel or batch normalization layer)

在跨通道或分批归一化的情况下，在激活层的过滤器区域内找到归一化值，并且通道数由用户确定，它是超级参数。对于 AlexNet，每个激活元素使用 5 个通道来估计归一化值(如图 4-3 所示)。这将有助于已经活跃的神经元的进一步兴奋，并可以检测到高频特征。

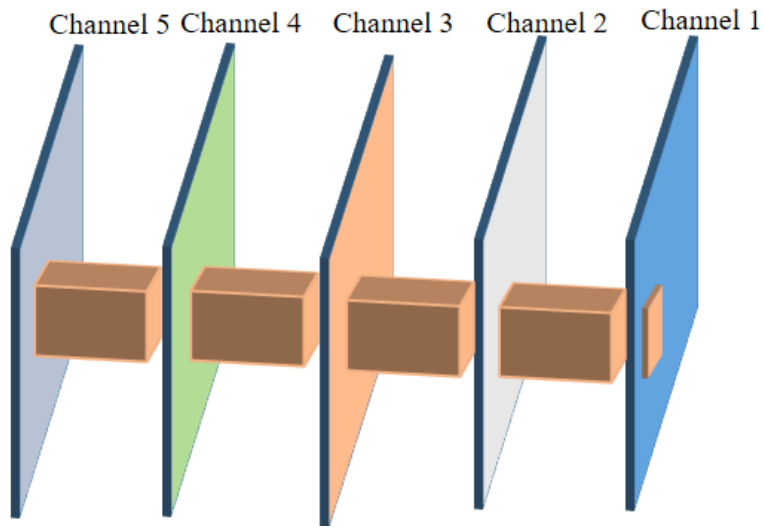

图 4-3 每个元素有 5 个通道的批处理规范化

5、全连接层(Fully connected layer)

全连接层类似于典型的人工神经网络,所有神经元相互连接。 AlexNet 体系结构由具有 4096 个神经元的两个完全连接的层组成(如图 4-4 所示)。最后的完全连接层已被 4 个神经元修饰,这对应于房价的大小。

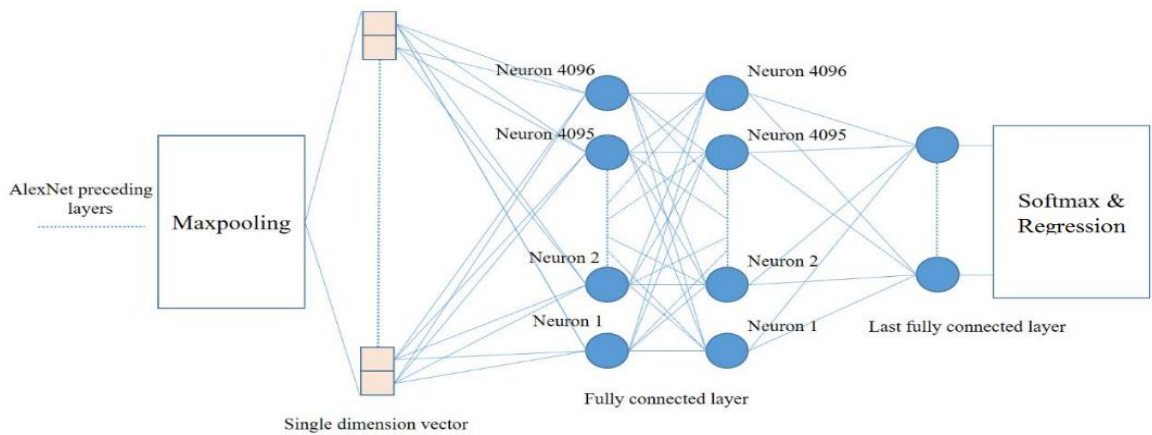

图 4-4 全连接层

6、脱落层(Dropout layer)和回归层

在脱落的情况下,神经元之间的连接会通过作为超参数的概率值来设置随机断开。对于 AlexNet,概率值为 0.5,这意味着每次运行随机断开 50％的神经元,这样可以提高通用性并减少依赖性。回归层可以直接输出回归层的大小。

## 4.2 预测评价指标

### 4.2.1 平均绝对值误差——MAE

MAE 是最基础的评估方法之一，代表了目标值与预测值之间差的绝对值的和。取值范围为$[0,+\infty)$，MAE 的值越小，说明预测模型拥有更好的精确度。相反，如果 MAE 的值越大，则误差越大。

$$MAE = \frac{1}{m}\sum_{i=1}^{m}|(y_i - \hat{y}_i)| \tag{1}$$

其中$y_i$为实际房价，$\hat{y}_i$为预测房价，下同。

### 4.2.2 均方误差损失函数——MSE

均方误差损失函数，通常是回归问题的首选损失函数，很多机器学习框架中的回归分析模块，都默认使用 MSE。误差平方化将会使较大的误差被放大，意味着模型主要针对大误差进行惩罚。范围$[0,+\infty)$，当预测值与真实值完全吻合时等于 0，即完美模型；误差越大，该值越大。

$$MSE = \frac{1}{N}\sum_{i=1}^{N}(y_i - \hat{y}_i)^2 \tag{2}$$

### 4.2.3 平均绝对百分比误差损失函数——MAPE

MAPE作为一个预测准确性的衡量指标，不仅仅考虑了预测值与真实值得误差，也考虑了误差与真实值之间的比例。取值范围为$[0,+\infty)$，MAPE 为 0%表示完美模型，MAPE 大于 100 %则表示劣质模型。

$$MAPE = \frac{100\%}{n}\sum_{i=1}^{n}\left|\frac{\hat{y}_i - y_i}{y_i}\right| \tag{3}$$

## 4.3 模型与方法

### 4.3.1 问题阐述

房价预测问题是一个回归问题。对于 $X$ 是给定的房屋数据（视觉图像和文本特征），$Y$ 是相关的房价。我们的最终目标是通过给定的 $X$ 来预测房价($Y'$)，是其最大程度的接近于实际房价 $Y$，即最小化 $Y'$和 $Y$ 之间的差异。

### 4.3.2 深度视觉与文本特征混合模型(MVTs)

我们提出的深度视觉与文本特征混合房价预测模型(简称 MVTs)的架构如图 4-5 所示，模型中有三个模块，分别为：深度特征模块、浅层 CNN 特征模块和文本特征模块。首先，我们从预先训练的 ResNet50 模型中提取深层特征。其次，设计了浅层 CNN 模块，通过三个重复模块直接从原始图像中提取浅层特征。第三，我们添加了文本特征，包括卧室数量、浴室数量、面积、邮政编码、构建年份、修缮年限、房屋类型、停车位和采光率，金融因素包括 GDP，M2、地区生产总值、CPI、PPI，人口因素包括各地区年末人口总数。最后，我们将这三个模块连接在一起，形成深度视觉与文本特征混合房价预测(MVTs)模型。该模型后由两个不同的损失函数组成，分别是平均绝对误差损失百分比(MAPE)和绝对平均误差 (MAE)。平均绝对百分比误差损失衡量预测价格与实际房价的平均百分比差异，绝对平均差异损失确保预测价格的平均值接近平均房价。

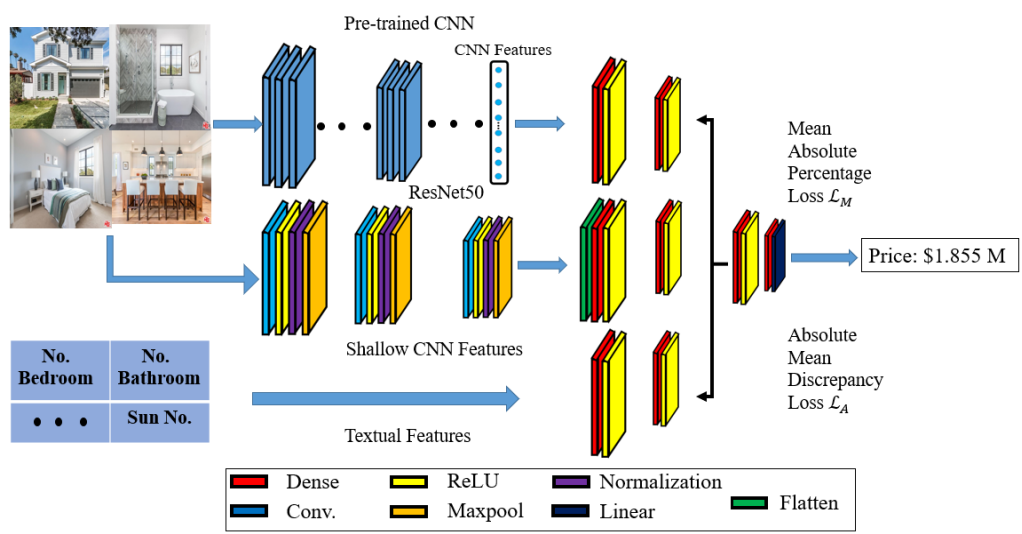

图 4-5 深度视觉与文本特征混合的房价预测模型(MVTs)

在深度特征模块中，我们遵循协议[14][27]，从最后一个完全连接层提取所有特征。因此，一个图像的最终输出为一个 1×1000 的向量。特征提取分为两个步骤：1、将图像进行预训练的神经网络重新调节为不同的输入尺寸；2、从最后一个完全连通层提取特征。在特征提取之后，接下来两个重复的模块。在每个模块中，有一个稠密层和一个"ReLu"激活层。全链接层的单位分别为 1000 和 4。

在浅层 CNN 特征模块中，包括三个重复模块，每个模块有四层：卷积二维层、"ReLu"激活层、批量归一化层和最大池(max pooling)层。卷积二维层中三个模块的

核大小为(3，3)，滤波器大小分别为 16，32，64，最大池(max pooling)层的池大小是(2，2)。在平坦层之后，是与深部特征模块相同的模块，但稠密层的单位是 16 和 4。

最后是文本特征模块，输入包括影响房价的因素，它是与深部特征模块相同的模块，但因为比其他两个模块具有更少的特征，所以稠密层的单元为 8 和 4。

接着将这三个模块连接在一起形成一个整体，该整体主要由四层组成：单元数为 4 的稠密层、"ReLu"激活层、单元数为 1 的稠密层，最后"线性"激活层结尾。因此，它可以在最后一层输出房价的预测结果。

最后我们的模型要最小化下面的目标函数：

$$L(x,y) = argmin\ L_M(x,y) + \alpha L_A(x,y) \tag{4}$$

$$L_M = \frac{1}{N}\sum_{i=1}^{N}\left|\frac{y_i - y_i'}{y_i}\right|,\ L_A = \left|\frac{\bar{y} - \bar{y}'}{\bar{y}}\right| \tag{5}$$

其中 $L_M$ 是平均绝对误差损失百分比(MAPE)，$L_A$ 是提出的绝对平均误差损失函数(MAE)，$\alpha$ 是两个损失函数之间的平衡因子，$\bar{y}$ 是平均房价。

本文提出的深度视觉与文本特征混合的房价预测模型(MVTs)有三个明显的优势。首先，我们从深层视觉特征和文本特征两方面考虑混合特征，考虑了房价更多的影响因素。第二，我们提出的深度视觉与文本特征混合的房价预测模型(MVTs)通过连接深度特征、浅层 CNN 特征和文本特征这三个模块，充分利用所有这些模块中的特性进行实验并获得更高性能。并且提出了一种新的损失函数作为目标函数，目的就是使预测的值更接近真实值。有效地度量了预测房价与实际房价的平均差。此外，所设计的稠密和激活模块能够成功地利用深层的视觉和文本特征来预测房价，且不会出现过拟合问题。如果能够添加更多的层（如，多次重复稠密层、ReLu 层、标准化层和脱落层），最终的误差将首先停留在全局最小，然后逐步增加。

# 5 模型的预测结果

## 5.1 国内整体房价预测结果

我们将国内四个城市房屋数据集随机分为训练数据和测试数据，其中每组数据包含四张房屋视觉图像和房屋文本特征，选取75%的数据进行训练，而后用剩余的25%进行测试。图 5-1 将预测所得房价与实际房价进行比较。我们发现模型预测出的房价与实际房价接近，这代表了我们提出的深度视觉与文本特征混合的房价预测模型(MVTs)适合于房价预测。

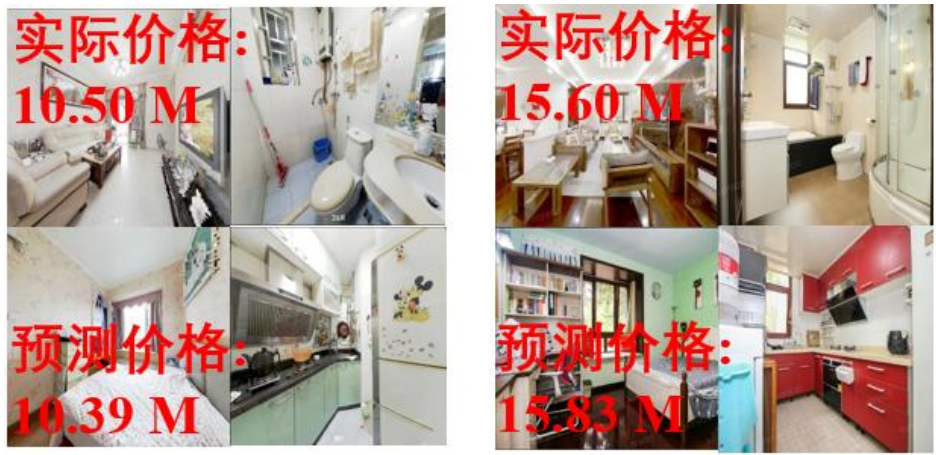

图 5-1 房价预测的两个例子。实际价格和预测价格在房屋图像中标注(M：百万)

为了说明我们提出的深度视觉与文本特征混合的房价预测模型(MVTs)对房价预测的有效性，我们得出了训练样本的平均绝对误差损失百分比(MAPE)和均方误差(MSE)。MAPE 在公式 3 中定义，MSE 由公式 2 定义。这两个指标的数值越小，该模型的性能越好。

表 5-1 列出了模型不同设置下的 MAPE 和 MSE 的值。我们首先在测试数据中分别对深度特征，文本特征和浅层 CNN 特征运行来预测房价，然后组合了这三个特征的进行实验结果。在所有三个不同的特性中，文本特性实现了比其他两个特性更高的性能，影响程度更大。尽管文本特征比深层特征具有的特征少，但它包含的信息比简单的房屋外部和内部(如位置和面积)更有用。此外，深度特征比浅层 CNN 特征有更好的效果，因为深度神经网络是由数百万的图像训练而成的。因此，它可以提取出优于浅 CNN 特征的特征。但这三个特征的组合实现了最高的性能，即最低的 MAPE 和 MSE。我们还观察到，所有四种不同的预测结果都比以前的 SVR 模型好。同时，我们也对比了其他 4 种上文中提到的常用的模型：BP，GM(1,1)，ANN

和 ARIMA。我们可以发现 ARIMA 模型的结果比 BP，GM 和 ANN 结果好，但是比我们提出的房价预测模型(MVTs)的结果误差更大。因此我们提出新的 MVTs 模型在所有模型中是最适用于房价预测的。

表 5-1 不同模型的房价预测结果（国内总体数据）

|  | 深度特征 | 文字特征 | 浅层CNN特征 | **MVTs** | SVR | BP | GM | ANN | ARIMA |
|---|---|---|---|---|---|---|---|---|---|
| MAPE(%) | 56.78 | 42.94 | 64.96 | **20.98** | 80.87 | 79.35 | 66.32 | 72.42 | 59.87 |
| MSE | 0.1446 | 0.1037 | 0.278 | **0.0874** | 0.3987 | 0.3874 | 0.2814 | 0.3601 | 0.1732 |

此外，MAPE 是比 MSE 更好的度量手段。对于浅层 CNN 和深度 CNN 特征，标准化均方误差比较接近，但两者的 MAPE 评分有显著差异。为进一步验证这一观察结果，我们在表 5-2 中展示了两个度量手段的有效性。我们发现，如果数据被标准化处理，MSE 的值会发生显著变化。然而，MAPE 的值仍然和以前一样，如表 5-2 所示，列出了不同数据处理的迭代次数和时间。这表明，与未规范化的数据相比，规范化的房价会导致更少的时间和迭代次数和更高的性能。因此，我们提出的 MVTs 模式在对标准化后的数据集进行的房价预测的效果更好。

表 5-2 不同数据处理的迭代次数和时间

|  | 迭代次数 | 时间(s) | MAPE | MSE |
|---|---|---|---|---|
| 非标准化价格 | 300 | 5096 | 36.90 | 0.1205 |
| 标准化价格 | **60** | **1080** | **20.98** | **0.0874** |

为了对比不同城市的结果，我们分别对每个城市的 1000 个数据进行了训练和测试。我们同样选取 75%的数据进行训练，选取 25%的样本数据进行预测。

## 5.2 北京房价预测结果

从 2017 年至 2020 年,北京的地区生产总值从 28014.94 亿元增加至 36102.6 亿元，年末人口总数从 2171 万降低至 2154 万左右。北京 2017 年平均的房价为 57768 元/m$^2$，2018 年 59868 元/m$^2$ 达到小规模巅峰，2019 至 2020 年受新冠疫情影响略有下降，但仍位居全国第一。

通过对收集到的北京的 1000 套房屋数据进行分析，从表 5-3 中可以看出，我们提出的深度视觉与文本特征混合的房价预测(MVTs)模型的 MAPE 和 MSE

值最小。对比前三列的 MAPE 和 MSE 值,我们发现如果仅用文字属性,深度或浅层图像特征进行预测,效果并不好。此外我们还对比了其他 5 种上文中提到的常用于房价预测的模型,我们的模型误差值分别为 18.26%和 7.25%,明显小于其他模型,证明了模型的有效性。

表 5-3 北京房价预测结果

|         | 深度特征 | 文字特征 | 浅层 CNN 特征 | **MVTs** | SVR | BP | GM | ANN | ARIMA |
|---------|---------|---------|--------------|----------|------|------|------|------|-------|
| MAPE(%) | 54.19   | 40.1    | 62.18        | **18.26** | 76.29 | 76.45 | 64.19 | 69.01 | 55.07 |
| MSE     | 0.1381  | 0.093   | 0.2196       | **0.0725** | 0.341 | 0.3596 | 0.2285 | 0.3013 | 0.1183 |

## 5.3 上海房价预测结果

上海作为我国最大的经济中心城市,地区生产总值位居全国第一,2017 至 2020 年增长 8067.59 亿元,年末人口总数增长 10 万人,近三年的房价起起伏伏,到 2020 年末,上海得平均房价为 52530 元/m²,位居全国第三。

通过对收集到的上海 1000 套房屋数据进行分析,可以发现,在上海房价预测中新提出的深度视觉与文本特征混合的房价预测模型(MVTs)仍然是最好的。此外,在 5 种传统模型中,结果优越性的顺序为:ARIMA > GM (1,1)> ANN > SVR > BP。

表 5-4 上海房价预测结果

|         | 深度特征 | 文字特征 | 浅层 CNN 特征 | **MVTs** | SVR | BP | GM | ANN | ARIMA |
|---------|---------|---------|--------------|----------|------|------|------|------|-------|
| MAPE(%) | 53.38   | 39.83   | 60.28        | **17.10** | 74.39 | 75.85 | 62.85 | 67.56 | 54.58 |
| MSE     | 0.1204  | 0.0882  | 0.1946       | **0.0680** | 0.3296 | 0.3347 | 0.2035 | 0.2953 | 0.1023 |

## 5.4 深圳房价的预测结果

深圳是全国创新能力第一的城市,经济总量仅次于上海、北京。近几年也吸引着大量的年轻人去往深圳创业和工作,三年的时间里年末人口总数增长 91.05 万人。商品房二级市场平均交易价格截止至 2019 年末,从 2017 年的 54430 元/m² 上涨至 55250 元/m²。

在对深圳的 1000 套数据进行房价预测时,我们仍然可以得到和表 5-1、3、4 相同的结论:我们提出的深度视觉与文本特征混合的房价预测模型(MVTs)是预测效果

最好的模型，与 5 种传统模型结果对比顺序是：ARIMA＞GM＞ANN＞SVR＞BP。

表 5-5 深圳房价预测结果

|  | 深度特征 | 文字特征 | 浅层 CNN 特征 | **MVTs** | SVR | BP | GM | ANN | ARIMA |
|---|---|---|---|---|---|---|---|---|---|
| MAPE(%) | 50.37 | 36.89 | 57.84 | **15.38** | 70.73 | 72.17 | 59.36 | 64.84 | 51.34 |
| MSE | 0.1012 | 0.0798 | 0.1734 | **0.0582** | 0.2903 | 0.3191 | 0.1893 | 0.2693 | 0.1102 |

## 5.5 广州房价的预测结果

广州是国务院定位的国际大都市，是中国第五个晋升 Alpha 级水平的城市，与北京、上海并称"北上广"。作为广东省的省会城市，占据着地理和经济环境的优势，也同样吸引着大量的人才落户生根。作为国内具有代表性的城市之一，它的房地产市场也是具有代表性的。

对收集到的数据进行实验，验证提出的深度视觉与文本特征混合的房价预测模型(MVTs)的有效性，我们仍可以发现，提出的模型的 MAPE 值和 MSE 值分别为 16.96%和 0.063，远低于其他模型，其他 5 种传统模型中结果优越性的顺序依旧：ARIMA ＞ GM ＞ ANN ＞ SVR ＞ BP。

表 5-6 广州房价预测结果

|  | 深度特征 | 文字特征 | 浅层 CNN 特征 | **MVTs** | SVR | BP | GM | ANN | ARIMA |
|---|---|---|---|---|---|---|---|---|---|
| MAPE(%) | 52.68 | 38.86 | 59.73 | **16.96** | 72.67 | 74.98 | 60.86 | 66.98 | 53.12 |
| MSE | 0.1196 | 0.0829 | 0.1846 | **0.063** | 0.3107 | 0.3848 | 0.1973 | 0.2836 | 0.1203 |

从表 5-3 到表 5-6，我们可以发现每个城市的预测结果都不同因为数据不同，而且房价的区间也不一样。在我们的深度视觉与文本特征混合的房价预测模型(MVTs)中，一个有趣的结果是，房价预测结果的优越性为：深圳 ＞ 广州 ＞ 上海 ＞ 北京。这个结果是由于每个城市房价不同造成的，而且也和收集数据的方差有关。我们进一步研究数据发现方差的顺序为：北京＞ 上海＞ 广州 ＞ 深圳。因此深度视觉与文本特征混合的房价预测模型(MVTs)显示了深圳房价预测的结果最好。

## 5.6 模型适用性评估

为了论证模型的性能，我们选取了国外的数据来论证。同样的在模型中，选取提取到的 75%的数据进行训练，选取 25%的样本数据进行拟合和预测。结果如下表所示，提出的深度视觉与文本特征混合的房价预测模型(MVTs)的预测结果领先其他

模型。证明了我们的模型，不论是在国内还是国外，预测精度都相对较高。

表 5-7 国外房价预测结果

|  | 深度特征 | 文字特征 | 浅层 CNN 特征 | **MVTs** | SVR | BP | ANN | ARIMA |
|---|---|---|---|---|---|---|---|---|
| MAPE(%) | 33.71 | 24.19 | 33.65 | **16.61** | 26.63 | 24.17 | 46.24 | 27.21 |
| MSE | 0.0032 | 0.0062 | 0.0052 | **0.0024** | 0.1304 | 0.009 | 0.0035 | 0.0043 |

为了证明模型的适用性，我们还使用之前的数据集测试了我们的模型。房屋数据来自 Ahmed[2]，它包含 535 幅房屋图像，我们使用 75%作为训练，25%作为测试。实验结果表明，我们的模型明显优于 SVR 模型。

表 5-8 使用 Ahmed 文中数据与 SVR 模型的比较

|  | **MVTs** | SVR |
|---|---|---|
| MAPE(%) | **34.29** | 56.09 |
| MSE(norm.) | **0.01734** | 0.1494 |

从以上结果来看，我们提出的深度视觉与文本特征混合的房价预测模型(MVTs)能够在预测房价时产生较低的 MAPE 和 MSE，这证明了我们的模型的稳健性。我们还比较了两种不同损失函数的有效性，如表 5-9 所示。我们发现，由于 $L_A$ 只测量绝对百分比误差的平均值，导致单独使用 $L_A$ 损失函数度量效果不如单独使用 $L_M$ 损失函数。然而，这两个损失函数的组合却实现了最高的性能。

表 5-9 不同损失函数下的预测结果

|  | $L_M$ | $L_A$ | 组合 |
|---|---|---|---|
| MAPE(%) | 22.10 | 25.77 | **18.01** |
| MSE | 0.00447 | 0.00589 | **0.00477** |

此外，我们还探讨了不同的预训练模型对最终结果的影响。我们检测了 16 个训练成熟的模型。分别是 SqueezeNet[28]、Alexnet[29]、Googlenet[30]、Shufflenet[31]、Resnet18[32]、Vgg16[33]、Vgg19[33]、Mobilenetv2[34]、Nasnetmobile[35]、Resnet50[32]、Resnet101[32]、Densenet201[36]、Inceptionv3[37]、Xception[38]、Inceptionresnetv2[39]、Nasnetlarge[35]。如图 5-2 和表 5-7 所示，ResNet50 显而易见的获得了比其他模型更高的性能。因此，选用 ResNet50 作为预训练模型，来提取图像中的深度特征信息。

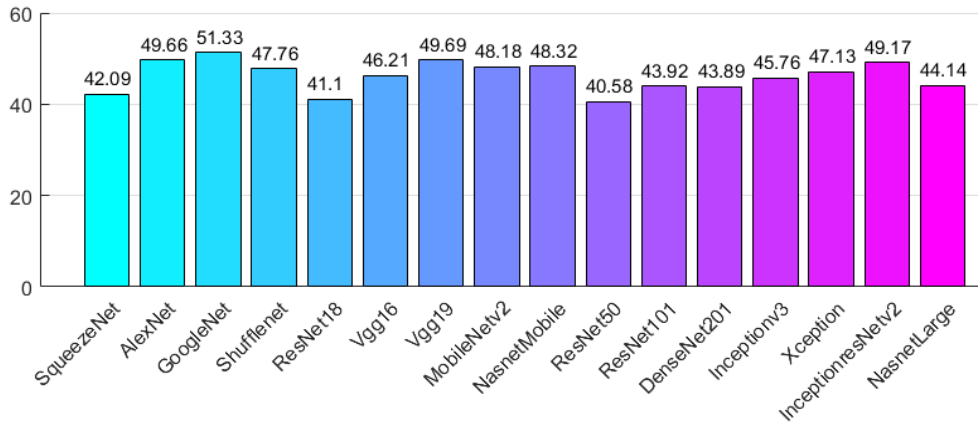

图 5-2 16 个预先训练的模型提取特征的 MAPE 值条形图(x 轴是 ImageNet 模型 top-1 正确率的阶数)

表 5-10 16 个预训练神经网络使用最小的 LM 和 L¬A 进行预测的 MAPE 和 MSE 的值

| 模型 | MAPE(%) | MSE |
| --- | --- | --- |
| SquezeNet | 42.09 | 0.01005 |
| Alexnet | 49.66 | 0.01095 |
| Googlenet | 51.33 | 0.00985 |
| Shufflenet | 47.76 | 0.00943 |
| Resnet18 | 41.10 | 0.00884 |
| Vgg16 | 46.21 | 0.00869 |
| Vgg19 | 49.69 | 0.00942 |
| Mobilenetv2 | 48.18 | 0.00946 |
| Nasnetmobile | 48.32 | 0.00925 |
| Resnet50 | **40.58** | **0.00652** |
| Resnet101 | 43.92 | 0.00794 |
| Densenet201 | 43.89 | 0.00830 |
| Inceptionv3 | 45.76 | 0.00818 |
| Xception | 47.13 | 0.00882 |
| Inceptionresnetv2 | 49.17 | 0.00851 |
| Nasnetlarge | 44.14 | 0.00869 |

我们又进一步探讨16个预先训练的神经网络的 *top-1* 正确率与图 5-3 中的 MAPE 值之间的关系。然而，我们发现相关系数和 $R^2$ 值相对较小，这些值越小，相关性越小[66]。因此，如何通过 *top-1* 正确率来选择最佳的预训练模型是一个不可忽视的趋势。虽然我们不知道其内在机制，但是 ResNet50 在房价预测问题中是有用的。

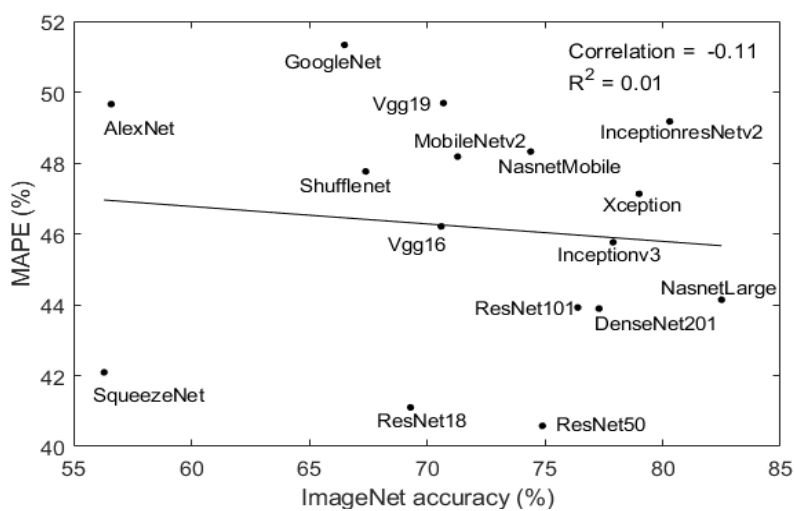

图 5-3 16 个预训练模型的 ImageNet 精度与 MAPE 在房价预测中的相关性

结果表明，图像特征(深度特征和浅层 CNN 特征)的性能分别都明显低于单一的文本特征。实际上，房价会受到很多因素的影响，比如文本特征中的因素，尤其是位置。然而，房子的视觉图像来自四种不同视角，它只考虑房子的外部和内部。因此，单一的深层特征或浅层 CNN 特征的性能比单一的文本特征差。

我们清楚地观察到我们提出的混合深层视觉和文本特征(MVTs)模型的几个优点。首先，我们的模型考虑了三个不同的特征：来自预训练模型的深层特征、浅层 CNN 特征和文本特征。其次，我们提出了一种新的损失函数，称为绝对平均差异损失，它可以联合降低预测房价和实际房价之间的差异。

实验显示新收集的数据的性能比之前数据集的结果要好。一是因为我们在训练阶段收集了更多的样本图像（视觉图像和文本数据），对一些样本进行训练，对比之前文章中的 538 个数据(其中 428 个数据进行训练，107 组进行实验)多出接近十倍的数量，训练数量更多，精度越大。二是我们收集了更多影响房价的因素。我们搜集到的新的房屋数据是 36 个月内的最新数据，根据地区的不同选取更多的文本影响因素进行实验，而 Ahmed[2]等人采用的是 2016 年左右的数据，仅用其中的 4 个文本影响因素进行实验。

## 5.7 经济和人口影响因素的重要性分析

上文中曾说道，除了房屋本身的影响因素之外，金融和人口因素同样对房价预测很重要。我们特别研究了经济和人口影响因素的重要性。因此，我们采用因子分析的方法去决定每个因素的重要性。如下图所示，地区生产总值、地区人口(population)，CPI 和 PPI 是四个最重要的经济和人口影响因素，因为他们的贡献率占比较大。

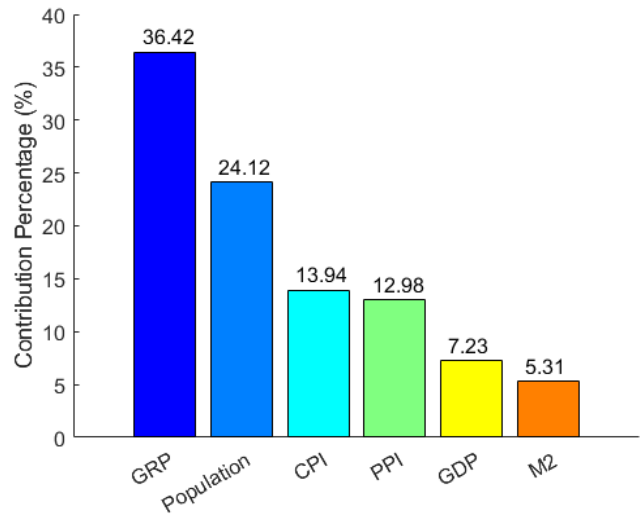

图 5-4 经济和人口影响因素贡献率

# 6 总结与展望

## 6.1 研究总结

在本文中，我们建立了一个将视觉和文字特征结合在在一个单元架构中的新模型来进行房价预测，该模型时具有较强的鲁棒性。模型中包含三个模块，分别为：深度特征模块、浅层 CNN 特征模块和文本特征模块。首先，我们从预先训练的 ResNet50 模型中提取深层特征。其次，设计了浅层 CNN 模块从原始图像中提取浅层特征。第三，我们添加了房价影响因素的文本特征。最后，我们将这三个模块连接在一起形成，用于房价预测。

通过实验，综合视觉图像和文本属性可以得到预测结果优于深度特征、浅 CNN 特征和文本特征。还与常被用于房价预测的模型，包括时间序列预测、GM(1,1)模型、支持向量回归(SVR)、BP 神经网络、人工神经网络的结果进行对比。结果显示，新提出的深度视觉与文本特征混合的房价预测模型(MVTs)的预测准确性更高。

## 6.2 未来展望

针对上述研究的结论和改进方向，结合当前的前沿思想和技术，这小节是对未来的相关方向进行思考和展望，希望能对后续的研究提供一些有益的参考。

房地产市场是我国国民经济命脉，也受到政府和相关部门的高度重视。但是，目前我国的房地产价格数据的主要来源大都是中介平台，这中间或许会存在不透明，有时也没有办法在海量的信息中筛选有用的加以利用，耗费大量人力和时间，且准确性无法度量。需要市场加强监管，完善相关制度，防止产生过高的泡沫。

随着研究的深入，作者认识到自己的研究仍有不足。首先对于数据集，我们提取到的房价预测的国内的数据集是最近三年的最新数据，但仅有北上广深四个城市共计 4000 套房屋样本数据，虽然最新，但远远不够为了提高后续实验的准确性，我们仍需提取更多的数据。其次，房价的影响因素也不仅仅是上述文章中所搜集到的文本特征，还有包括文化、政策，甚至是周边的配套设施情况，如学区，地铁等仍有待搜集。如果有相关的数据源，那我们的模型的准确度会更高。

至于模型也有一些局限性。首先，测试数据的最低 MAPE 较高，这意味着预测的房价在某种程度上偏离实际价格，虽远低于其他模型的结果，但仍有改进的空间。其次，两种视觉特征（深度特征和浅层 CNN 特征）的 MAPE 值都相对较低，说明图像的深层结构需要进一步研究。

# 参考文献

[1]  Wrenn D H , Yi J , Zhang B . House prices and marriage entry in China[J]. Regional Science and Urban Economics, 2018.

[2]  Eman Ahmed and Mohamed Moustafa. House price estimation from visual and textual features. arXiv preprint arXiv:1609.08399, 2016.

[3]  郑永坤, 刘春. 基于 ARIMA 模型的二手房价格预测[J]. 计算机与现代化, 2018.

[4]  Chih-Hung Wu, Chi-Hua Li, I-Ching Fang, Chin-Chia Hsu, Wei-Ting Lin, and Chia-Hsiang Wu. Hybrid genetic-based support vector regression with feng shui theory for appraising real estate price. In 2009 First Asian Conference on Intelli- gent Information and Database Systems, pages 295–300. IEEE, 2009.

[5]  Ian D Wilson, Stuart D Paris, J Andrew Ware, and David Harrison Jenkins. Resi- dential property price time series forecasting with neural networks. In Applications and Innovations in Intelligent Systems IX, pages 17–28. Springer, 2002.

[6]  高玉明, 张仁津. 基于遗传算法和 BP 神经网络的房价预测分析%Analysis of House Price Prediction Based on Genetic Algorithm and BP Neural Network[J]. 计算机工程, 040(004):187-191.

[7]  李东月. 房价预测模型的比较研究[J]. 工业技术经济, 2006, 025(009):65-67.

[8]  刘丽泽. 基于多元线性回归模型及 ARIMA 模型的北京市房价预测[J]. 科技经济导刊, 2018, 26(29):182-183.

[9]  武秀丽, 张锋. 时间序列分析法在房价预测中的应用——以广州市的数据为例[J]. 科学技术与工程, 2007(21):171-175.

[10] Azme Bin Khamis and Nur Khalidah Khalilah Binti Kamarudin. Comparative study on estimate house price using statistical and neural network model. Inter-national Journal of Scientific & Technology Research, 3(12):126–131, 2014.

[11] Visit Limsombunchai. House price prediction: hedonic price model vs. artificial neural network. In New Zealand Agricultural and Resource Economics Society Conference, pages 25–26, 2004.

[12] Ayush Varma, Abhijit Sarma, Sagar Doshi, and Rohini Nair. House price predic-tion using machine learning and neural networks. In 2018 Second International Conference on Inventive Communication and Computational Technologies (ICI- CCT), pages 1936–1939. IEEE, 2018.

[13] JJ Wang, SG Hu, XT Zhan, Q Luo, Qi Yu, Zhen Liu, Tu Pei Chen, Y Yin, Sumio Hosaka, and Yang Liu. Predicting house price with a memristor-based artificial neural network. IEEE Access, 6:16523–

16528, 2018.

[14] Youshan Zhang and Brian D Davison. Modified distribution alignment for domain adaptation with pre-trainedinception resnet. arXiv preprint arXiv:1904.02322, 2019.

[15] Youshan Zhang, Sihong Xie, and Brian D Davison. Transductive learning via improved geodesic sampling. In Proceedings of the 30th British Machine Vision Conference, 2019.

[16] Youshan Zhang, Jon-Patrick Allem, Jennifer Beth Unger, and Tess Boley Cruz. Automated identification of hookahs (waterpipes) on instagram: an application in feature extraction using convolutional neural network and support vector machine classification. Journal of medical Internet research, 20(11):e10513, 2018.

[17] Youshan Zhang and Brian D Davison. Impact of ImageNet Model Selection on Domain Adaptation. In *Proceedings of the IEEE winter conference on Applications of Computer Vision Workshops,* Pages 173-182, 2020.

[18] Youshan Zhang and Brian D Davison. Domain adaptation for object recognition using subspace sampling demons. *Multimedia Tools and Applications*: 1-20.

[19] Youshan Zhang and Brian D Davison. Adversarial Consistent Learning on Partial Domain Adaptation of PlantCLEF 2020 Challenge. In: *CLEF working notes2020, CLEF: Conference and Labs of the Evaluation Forum*, Sep. 2020, Thes-saloniki, Greece.

[20] Youshan Zhang, Hui Ye and Brian D Davison. Adversarial Reinforcement Learning for Unsupervised Domain Adaptation. In *Proceedings of the IEEE/CVF Winter Conference on Applications of Computer Vision*, pp. 635-644. 2021.

[21] Youshan Zhang and Brian D Davison. Adversarial Continuous Learning on Unsupervised Domain Adaptation. In *25th International Conference on Pattern Recognition Workshops*, pp. 672-687. 2021.

[22] Youshan Zhang and Brian D Davison. Adversarial Regression Learning for Bone Age Estimation. In *International Conference on Information Processing in Medical Imaging*, pp. 742-754. Springer, Cham, 2021.

[23] Youshan Zhang and Brian D Davison. Deep Spherical Manifold Gaussian Kernel for Unsupervised Domain Adaptation. In *Proceedings of the IEEE/CVF Conference on Computer Vision and Pattern Recognition Workshops*, pp.4443-4452. 2021.

[24] Youshan Zhang and Brian D Davison. Efficient Pre-trained Feature Selection and Recurrent Pseudo-Labeling in Domain Adaptation. In *Proceedings of the IEEE/CVF Conference on Computer Vision and Pattern Recognition Workshops*, pp. 2719-2728. 2021.

[25] Youshan Zhang and Brian D Davison. Correlated Adversarial Joint Discrepancy Adaptation Network for Unsupervised Domain Adaptation. In *2021 International Conference on Content-Based Multimedia Indexing*, 2021.

[26] Youshan Zhang and Brian D Davison. Enhanced Separable Disentanglement for Un-supervised Domain Adaptation. In *2021 IEEE International Conference on Image Processing*, 2021

[27] Youshan Zhang and Brian D Davison. Weighted Pseudo Labeling Refinement for Plant Identification. In: *CLEF working notes 2021, CLEF: Conference and Labs of the Evaluation Forum*, 2021.

[28] Forrest N Iandola, Song Han, Matthew W Moskewicz, Khalid Ashraf, William J Dally, and Kurt Keutzer. Squeezenet: Alexnet-level accuracy with 50x fewer pa- rameters and¡ 0.5 mb model size. arXiv preprint arXiv:1602.07360, 2016.

[29] Alex Krizhevsky, Ilya Sutskever, and Geoffrey E Hinton. Imagenet classification with deep convolutional neural networks. In Advances in Neural Information Pro- cessing Systems, pages 1097–1105, 2012.

[30] Christian Szegedy, Wei Liu, Yangqing Jia, Pierre Sermanet, Scott Reed, Dragomir Anguelov, Dumitru Erhan, Vincent Vanhoucke, and Andrew Rabinovich. Going deeper with convolutions. In Proceedings of the IEEE conference on computer vision and pattern recognition, pages 1–9, 2015.

[31] Xiangyu Zhang, Xinyu Zhou, Mengxiao Lin, and Jian Sun. Shufflenet: An extremely efficient convolutional neural network for mobile devices. In Proceedings of the IEEE Conference on Computer Vision and Pattern Recognition, pages 6848– 6856, 2018.

[32] Kaiming He, Xiangyu Zhang, Shaoqing Ren, and Jian Sun. Deep residual learning for image recognition. In Proceedings of the IEEE Conference on Computer Vision and Pattern Recognition (CVPR), pages 770–778, 2016.

[33] Karen Simonyan and Andrew Zisserman. Very deep convolutional networks for large-scale image recognition. arXiv preprint arXiv:1409.1556, 2014

[34] Mark Sandler, Andrew Howard, Menglong Zhu, Andrey Zhmoginov, and Liang Chieh Chen. Mobilenetv2: Inverted residuals and linear bottlenecks. In Proceedings of the IEEE Conference on Computer Vision and Pattern Recognition, pages 4510–4520, 2018.

[35] Barret Zoph, Vijay Vasudevan, Jonathon Shlens, and Quoc V Le. Learning trans- ferable architectures for scalable image recognition. In Proceedings of the IEEE conference on computer vision and pattern recognition, pages 8697–8710, 2018.

[36] Gao Huang, Zhuang Liu, Laurens Van Der Maaten, and Kilian Q Weinberger. Densely connected convolutional networks. In Proceedings of the IEEE conference on computer vision and pattern recognition, pages 4700–4708, 2017.

[37] Christian Szegedy, Vincent Vanhoucke, Sergey Ioffe, Jon Shlens, and Zbigniew Wojna. Rethinking the inception architecture for computer vision. In Proceedings of the IEEE conference on computer vision and pattern recognition, pages 2818– 2826, 2016.

[38] Francois Chollet. Xception: Deep learning with depthwise separable convolutions. In Proceedings of the IEEE Conference on Computer Vision and Pattern Recogni- tion, pages 1251–1258, 2017.

[39] Christian Szegedy, Sergey Ioffe, Vincent Vanhoucke, and Alexander A Alemi. Inception-v4, inception-resnet and the impact of residual connections on learning. In Thirty-First AAAI Conference on Artificial Intelligence, 2017.

[40] Youshan Zhang and Qi Li. A regressive convolution neural network and support vector regression model for electricity consumption forecasting. In Future of Information and Communication Conference, pages 33–45. Springer, 2019.

[41] Youshan Zhang, Liangdong Guo, Qi Li, and Junhui Li. Electricity consumption forecasting method based on MPSO-BP neural network model. Atlantis. Nov.2016.

[42] 欧阳建涛. 非线性灰色预测模型在房地产投资价格中的应用[J]. 工业技术经济, 2005, 024(005):78-80.

[43] 李东月，马智胜. 灰色 GM(1,1)模型在房价预测中的算法研究[J]. 企业经济, 2006, 000(009):96-98.

[44] 刘琼芳. 基于GM(1,1)模型的福州市房价预测[J]. 福建金融管理干部学院学报, 2018.

[45] 盛宝柱, 古铃. 基于 GM(1,1)模型的合肥市商品房房价预测[J]. 皖西学院学报, 2018, 34(05):47-51.

[46] 王莹, 王志祥. 基于灰色系统 GM(1,1)的淮安市房价预测模型[J]. 淮阴师范学院学报(自然科学版), 2017(1).

[47] 侯普光, 乔泽群. 基于小波分析和 ARMA 模型的房价预测研究[J]. 统计与决策, 2014, 000(015):20-23.

[48] 申瑞娜, 曹昶, 樊重俊. 基于主成分分析的支持向量机模型对上海房价的预测研究[J]. 数学的实践与认识, 2013(23):13-18.

[49] 谷秀娟, 李超. 基于马尔科夫链的房价预测研究[J]. 消费经济, 2012, 000(005):40-42,48.

[50] 韦光兰, 邓晓盈, 张琼. 基于马尔可夫链预测模型的昆明市房价预测分析[J]. 中国市场,

2015, 000(021):86-87,104.

[51] 顾莹, 夏乐天. 加权马尔可夫链在房价预测中的应用[J]. 重庆理工大学学报:自然科学版, 2013, 27(8):125-130.

[52] 张卉. 基于粒子群优化 BP 神经网络的房价预测[J]. 价值工程, 2012, 031(014):207-209.

[53] 闫妍, 徐伟, 部慧, 等. 基于 TEII@I 方法论的房价预测方法[J]. 系统工程理论与实践, 2007, 27(7):1-9.

[54] 邱启荣, 于婷. 基于主成分分析的 BP 神经网络对房价的预测研究[J]. 湖南文理学院学报(自然科学版), 2011(03):28-30+40.

[55] 王奕翔,陈济颖,王晟全,李昂. 基于改进型 RF-BP 神经网络的房地产价格预测[J]. 工业控制计算机. 2019(10)

[56] Yuying Wu and Youshan Zhang. Mixing Deep Visual and Textual Features for Image Regression. In *Proceedings of the 2020 Intelligent Systems Conference*. Volume 1. 2020.

[57] X Chen, W Lai, J Xu. House Price Prediction Using LSTM[C].2017

[58] Sommervoll D E , Sommervoll V . Learning from man or machine: Spatial aggregation and house price prediction[J]. CLTS Working Papers, 2019.

[59] Montero, Jose-Maria, Minguez, et al. Housing price prediction: parametric versus semi-parametric spatial hedonic models (vol 20, pg 27, 2017)[J]. Journal of Geographical Systems, 2018.

[60] Nur A , Ema R , Taufiq H , et al. Modeling House Price Prediction using Regression Analysis and Particle Swarm Optimization Case Study : Malang, East Java, Indonesia[J]. International Journal of Advanced Computer Science and Applications, 2017, 8(10).

[61] Gupta R , Marfatia H A , Pierdzioch C , et al. Machine Learning Predictions of Housing Market Synchronization across US States: The Role of Uncertainty[J]. The Journal of Real Estate Finance and Economics, 2021.

[62] B Yang, B Cao. Research on Ensemble Learning-based Housing Price Prediction Model[J]. Big Geospatial Data and Data Science(2018)1:1-8

[63] Youshan Zhang. Bayesian Estimation for Fast Sequential Diffeomorphic Image Variability. In *Science and Information Conference*, pp. 687-699. Springer, Cham, 2019.

[64] Serrano W. The Random Neural Network in Price Predictions[M]. 2020

[65] Kostic Z, Jevremovic A . What Image Features Boost Housing Market Predictions?[J]. IEEE Transactions on Multimedia, 2020, PP(99):1-1.

[66] Youshan Zhang and Brian D Davison. Shapenet: Age-focused landmark shape prediction with

regressive cnn. In 2019 International Conference on Content-Based Multimedia Indexing (CBMI), pages 1–6. IEEE, 2019.

[67] Hongfa Ding, Youliang Tian, Changgen Peng, Youshan Zhang, and Shuwen Xiang. Inference attacks on genomic privacy with an improved HMM and an RCNN model for unrelated individuals. *Information Sciences*, 512, pp.207-218, 2020.

[68] Kain J F , Quigley J M . Measuring the Value of Housing Quality[J]. Publications of the American Statistical Association, 1970, 65(330):532-548.

[69] Wabe, Stuart J . A Study of House Prices as a means of Establishing the Value of Journey Time, the Rate of Time Preference and the Valuation of some Aspects of Environment in the London Metropolitan Region[J]. Applied Economics, 1971, 3(4):247-255.

[70] Anderson R , Crocker T . Air Pollution and Residential Property Values[J]. Urban Studies, 1971, 8(3):171-180.

[71] Chin T L , Chau K W . A critical review of literature on the hedonic price model[J]. International Journal for Housing Science & Its Applications, 2003, 27(2):145-165.